\documentclass[twoside,11pt]{article}

\usepackage[submission]{melba}  
\usepackage{lipsum}		
\usepackage{makecell}

\usepackage{graphicx}

\pdfoutput=1
\usepackage{amsmath,amsfonts}
\usepackage{breqn}

\melbaheading{0}{https://www.melba-journal.org/article/XX-AA}{2020}{1-?}{mm/yyyy}{mm/yyyy}{Mac et al.}{}{}

\ShortHeadings{Adapting to Unseen Scanner Domains}{Mac et al.}
\firstpageno{1}

\title{Adapting to Unseen Vendor Domains for MRI Lesion Segmentation}

\author{\name Brandon Mac \email bmac@ryerson.ca \\  
	\addr Image Analysis in Medicine Lab (IAMLAB), Ryerson University, Toronto, ON, Canada
	\AND
	\name Alan R. Moody \email alan.moody@sunnybrook.ca \\
	\addr Department of Medical Imaging, University of Toronto, Toronto, ON, Canada \\
	\addr Department of Medical Imaging, Sunnybrook Health Sciences Centre, Toronto, ON, Canada
	\AND
	\name April Khademi \email akhademi@ryerson.ca \\
	\addr Image Analysis in Medicine Lab (IAMLAB), Ryerson University, Toronto, ON, Canada \\
	\addr Keenan Research Centre for Biomedical Science, St. Michael’s Hospital, Toronto, ON, Canada\\
	\addr Institute for Biomedical Engineering, Science, and Technology (iBEST), A Partnership Between St. Michael's Hospital and Ryerson University, Toronto, Canada
}

\begin{document}

\maketitle

\begin{abstract}
%

One of the key limitations in machine learning models is poor performance on data that is out of the domain of the training distribution. This is especially true for image analysis in magnetic resonance (MR) imaging, as variations in hardware and software create non-standard intensities, contrasts, and noise distributions across scanners. Recently, image translation models have been proposed to augment data across domains to create synthetic data points. In this paper, we investigate the application an unsupervised image translation model to augment MR images from a source dataset to a target dataset. Specifically, we want to evaluate how well these models can create synthetic data points representative of the target dataset through image translation, and to see if a segmentation model trained these synthetic data points would approach the performance of a model trained directly on the target dataset. We consider three configurations of augmentation between datasets consisting of translation between images, between scanner vendors, and from labels to images. It was found that the segmentation models trained on synthetic data from labels to images configuration yielded the closest performance to the segmentation model trained directly on the target dataset. The Dice coeffcient score per each target vendor (GE, Siemens, Philips) for training on synthetic data was 0.63, 0.64, and 0.58, compared to training directly on target dataset was 0.65, 0.72, and 0.61. Our code is available at~\url{https://github.com/IAMLAB-Ryerson/MRI-Augmentation}.


\end{abstract}


\section{Introduction}

Imaging of white matter hyper-intensities (WMH) plays a significant role in the management, diagnosis, and evaluation of treatment of multiple sclerosis and dementia \citep{polman_diagnostic_2011}. Manual delineations provide important volumetric information of lesion load and spatial distributions, however they are labor intensive to acquire and are often subject to high inter- and intra- rater variability \citep{egger_mri_2017, steenwijk_accurate_2013}. As a result, automated lesion segmentation methods are highly sought after due to their potential to streamline clinical workflows. 

In recent years, deep learning has becoming increasingly popular for medical image processing and analysis. Deep convolutional neural networks (CNNs) leverage the capacity to learn strong representations from a large amount of annotated data. However, large annotated datasets of medical images are often difficult to obtain, especially for tasks such as semantic segmentation where exhaustive  pixel-wise annotations are required. Furthermore, models often perform poorly when tested on populations from outside the distribution of the training set.  This is commonly referred to as domain shift problem \citep{quinonero-candela_covariate_2008}. 

Domain shift is especially problematic for MR images as variations in hardware and software create non-standard intensities, contrasts, and noise distributions, which is also referred to as the multi-center (MC) effect \citep{reiche_pathology-preserving_2019}. The MC effect is known to create different image characteristics between scanners and centers even for the same patient and same tissues  \citep{reiche_pathology-preserving_2019, zhong_automatic_2012}. To address domain shift, many authors have suggested utilizing  domain adaptation techniques. In general, domain adaptation aims to transfer knowledge from a source domain to a target domain, typically by leveraging domain invariant features to be translated \citep{volpi_generalizing_2018,sundaresan_comparison_2021}. There has been success found in various applications where a small labeled dataset was able to incorporate larger unlabeled datasets by domain adaption, which allowed for greater generalizability \citep{ganin_domain-adversarial_2016,wilson_survey_2020}. While domain adaption typically seeks to learn invariant features from real data, there have been recent developments in the application of generative adversarial networks (GAN) to create synthetic data. The objective of GANs is to generate synthetic data representative of a target distribution. This is achieved through adversarial training of a pair of generator and discriminator models, in which the generator captures the data distribution, and the discriminator estimates the probability that the data is real or generated \citep{goodfellow_generative_2014}. The capacity to generate synthetic data points have lead authors to explore the prospect of utilizing these methods to augment limited data sets \citep{sandfort_data_2019, frid-adar_synthetic_2018, huo_synseg-net_2019}. 

A promising trend of medical applications building upon CycleGAN \citep{zhu_unpaired_2020} have emerged for cross-modal synthesis from unpaired training images. In \cite{zhang_translating_2019} and \cite{huo_synseg-net_2019}, the authors have demonstrated the capacity to utilize cycle-based methods to augment between MR and CT images for both data synthesis and segmentation. By having a pair of generator and discriminator networks learn the respective distributions of each domain, synthetic data points can be generated through image translation from one domain to another. The experimental results from these works have shown the capacity for the translation to be annotation-preserving in the application of semantic segmentation. Building upon this, rather than the application of cross-domain translation between modalities, in this paper we investigate the application of  image translation models to MR images between scanner domains to address the MC problem. 
\subsection{Contribution}

\begin{figure}[!htbp]
\centering
\includegraphics[width=0.8\linewidth]{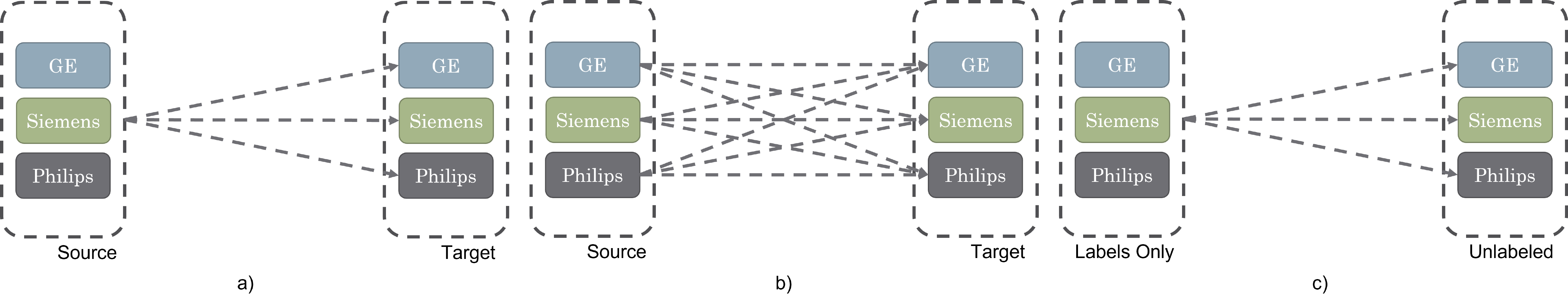}
\caption{Image-translation modes to be investigated. a) Image2Image translation, b) Scan2Scan, c) Label2Image}
\label{fig:domains}
\end{figure}

In this work, image translation techniques are utilized to augment MR images from one scanner domain to another for the purposes of semantic segmentation. Specifically, we seek to evaluate the capacity of these techniques to translate images from a source domain to create synthetic data points representative of the target domain. Rather than generalization, the objective is to see how well these images represent the target domain, and if training a segmentation model to learn from these synthetic images is comparable to a segmentation model learning directly from the target domain. We take inspiration from the work of \cite{zhang_translating_2019} and \cite{huo_synseg-net_2019} for cross-modal synthesis between MR and CT modalities, however apply these techniques to cross-domain translation between MR scanners. To elaborate, our investigation consists of two multi-center MR image datasets, from which we designate one as our source dataset and the other as our target dataset. We further separate the target dataset by the respective vendors (GE, Siemens, and Philips) and consider each as individual target distribution. For our experiments, we consider the following cases to be explored for defining the source distribution:

\begin{itemize}
    \item Image2Image - Translation from multi-center distribution to target distribution
    \item Scan2Scan - Vendor specific translation from source to target distribution
    \item Label2Image - Unpaired label to image translation from source to target distribution
\end{itemize}

Figure \ref{fig:domains} provides an illustration of the image translation cases. The overall objective was to define the optimal configuration to generate synthetic samples in a target domain through image translation. Image2Image utilizes the entire source dataset to train a translation model to each of the respective target distributions. This was to observe if the multi-center effect would translate to a specific target distribution. Contrasting this, Scan2Scan separates the source dataset by the respective vendors and trains a translation model per specific combination of source-target vendor pairs (i.e. GE2GE, Siemens2Philips, and Philips2GE). Lastly in Label2Image, we investigate defining the source distribution as the label images from the source dataset, and explore the potential of training a translation model for unpaired segmentation annotations to images in each target domain. From these models, we then consider two sets of experiments for training the segmentation models on synthetic data to further explore the multi-center effect. We define the synthetic datasets per method as the following: 

\begin{itemize}
    \item Generate synthetic data representative of a single target distribution
    \item Pool the synthetic data from each target distribution together to create a synthetic multi-center dataset
\end{itemize}

Our investigations found that the multi-center effect had little change on the quality of the synthetic images for image-to-image translation (Image2Image and Scan2Scan), however, did impact the overall variance of the segmentation performance. It was found that Label2Image mode yielded the best results for generating synthetic data for segmentation training, with results comparable to training directly on the target distribution. Interestingly, while the Label2Image synthetic data did train segmentation models better, it was noted that Label2Image yielded images that were relatively less realistic than the other modes. To explore the relationship between realism and effective training data further, we also conduct an experiment generating synthetic source labels using DCGAN \citep{radford_unsupervised_2016} to translate to the respective target distributions. 

\section{Related Works}
This work involves various aspects of medical image analysis, domain adaption, and generative adaptive networks (GANs). In the following sections, an overview of the relevant works is presented. 

\subsection{Semantic Segmentation}
Recent advances in semantic segmentation have largely been dominated by deep learning methodologies \citep{khademi_segmentation_2021}. Notably, the most common methods are based on fully convolutional networks (FCNs). \cite{long_fully_2015} demonstrated one can effectively transform any CNN model to an FCN. In medical applications, the most common FCN is encoder-decoder style U-Net first proposed by \cite{ronneberger_u-net_2015}. Exemplifying this, the top 11 teams participating in the MICCAI 2017 Grand Challenge for automated WMH segmentation used some form of U-Net architecture \citep{kuijf_standardized_2019}. However, it is usually expensive and difficult to acquire pixel-level annotations. Coupled with this, it is known that these models perform well on populations sampled from the training distribution, but perform poorly on out-of-domain populations \citep{volpi_generalizing_2018}. To address this, authors have proposed solutions such as weak supervision, where weak labels such as bounding boxes \citep{khoreva_simple_2016}, object localization \citep{leibe_augmented_2016}, and point supervision \cite{bearman_whats_2016}, to then refine to pixel-wise predictions. However, weak supervision does not perform as well as supervised methods as it is difficult to approximate boundary information from weak labels \citep{hung_adversarial_2018}.   
\subsection{Domain Adaptation}

To address the issues above, various domain adaption methods have been proposed to incorporate unlabeled data with labeled datasets. In general, domain adaption methods attempt to transfer knowledge from a source domain to a target domain by leveraging domain invariant features \citep{wilson_survey_2020}. A common technique is transfer learning, in which low level features are frozen, while higher level features are fine tuned \citep{ghafoorian_transfer_2017}. However, the performance is often known to rely on sufficient initial training data to be robust. Another domain adaption model is DANN, which consists of a feature extractor, a domain predictor and a label predictor \citep{ganin_domain-adversarial_2016}. The adversarial nature of the network is the gradient reversing layer between the domain predictor and the feature extractor, in which the labels prediction is minimized, while the domain predictor maximized. Similarly, an alternative is presented in domain unlearning, in which the method involves learning the domain prediction for a fixed feature representation, and then minimizing the domain shift, while maximizing in domain confusion \citep{dinsdale_deep_2021}.  
\subsection{Generative Adversarial Networks}
Recent developments in generative adversarial networks (GANs) \citep{goodfellow_generative_2014} have shown a variety of applications. As mentioned earlier, adversarial networks have been applied to domain adaptation \citep{ganin_domain-adversarial_2016}. \cite{hung_adversarial_2018} demonstrate the application of adversarial networks for semi-supervised training of semantic segmentation models. For data generation, application in the medical domain has been demonstrated by \cite{frid-adar_synthetic_2018} to generate synthetic liver images to improve lesion classification. In \cite{li_semantic_2021}, the authors propose a network to model the joint image-label distribution, and synthesis of both image and semantic mask. Similarly, \cite{zhang_datasetgan_2021} proposes training on a small labeled set and disentangling the latent space of StyleGAN \citep{karras_style-based_2019} architecture to generate synthetic labeled datasets. However, the authors note that the requirement is highly detailed semantic labels. \\
\indent Several authors have proposed utilizing CycleGAN \citep{zhu_unpaired_2020} as a method to adapt images from one domain to another for the purpose of augmenting training data. Translation from MR to CT domains have been demonstrated by several authors \citep{wolterink_deep_2017, zhang_translating_2019, jiang_tumor-aware_2018}. The original authors of CycleGAN have demonstrated the capacity to synthesize realistic images from label information \citep{zhu_unpaired_2020}, in a similar fashion to conditional GANs \citep{isola_image--image_2018}. A similar work to our investigation, \cite{palladino_unsupervised_2020} utilize CycleGANs to translate between scanner vendors within the MICCAI 2017 WML Grand Challenge data \citep{kuijf_standardized_2019}. However, in their implementation, they train only on real data and use the image translation models to augment the images the training distribution during testing.

\section{Methods}

\begin{figure}[!t]
\centering
\includegraphics[width=0.8\linewidth]{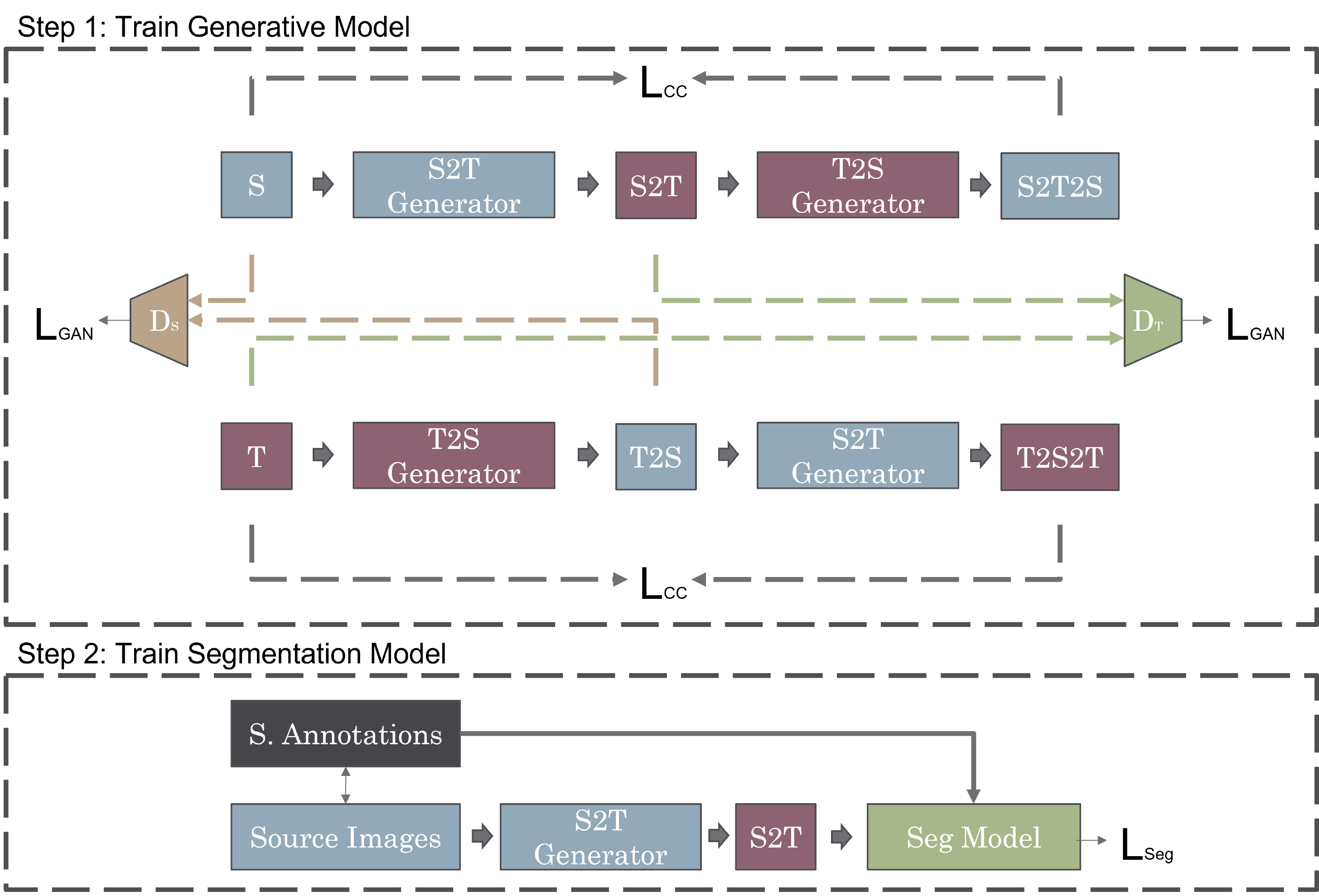}
\caption{Outline of Training Setup}
\label{fig:training_setup}
\end{figure}

An overview of the training setup is shown in Figure \ref{fig:training_setup}. Details regarding pre-processing, data, architecture and evaluation metric can be found below. 

\subsection{Pre-Processing}

Brain volumes were first skull stripped using method proposed by \cite{digregorio_intracranial_2021} to estimate the intracranial volume. Masks generated by the mentioned method were then used as a label along with WML labels for the source dataset. After stripping the skull, the whole volume was normalized to an intensity between 0 and 1 by dividing by maximum intensity. From the volume, axial slices were taken and resampled to 256 x 256 using default resize parameters in python scikit-image library \citep{van_der_walt_scikit-image_2014}. Slices that contained brain images with pixels that have less than 10\% of the volume or were empty were discarded.  This resulted in the upper- and lower-most slices being discarded, which were considered acceptable as the those regions were not expected to contain lesions (and had hardly any brain tissue) \citep{wardlaw_what_2015}. This was done to ensure that when training the translation models that they do not mode collapse to empty slices. Some moderate augmentations were applied to the images during training, including random scaling, rotation, and mirroring. 

\subsection{Data}

Two multi-center MR image datasets were used in our experiments, from which we designate one as our source dataset and the other as our target dataset. The source data consists of 60 FLAIR volumes publicly released by the MICCAI 2017 Grand Challenge for WMH segmentation \citep{kuijf_standardized_2019}. We designate the MICCAI dataset as the source dataset, as the annotations are considered the gold standard due to the multiple expert raters per annotations. The target dataset consists of 54 FLAIR volumes from the Canadian Atherosclerosis Imaging Network (CAIN) and is a pan-Canadian effort to study vascular disease \citep{tardif_atherosclerosis_2013}, with WML labels generated by experienced raters \citep{khademi_segmentation_2021}. The labels belonging to CAIN were not used in any training, and were strictly reserved for validation. 

A summary of the total number of images after pre-processing for both source and target distributions can be found in Table \ref{table:data}. In the table, the data is split into the respective scanner vendors. For training the image translation models, we pool the data as per each mode defined above in Figure \ref{fig:domains}. Regarding Image2Image and Label2Image, all images and labels (respective of each mode) from the source (MICCAI) distribution were pooled together as a set containing 1,695 images. From this set, individual translation models were trained to translate to each of the vendors in the target distribution (341 images in GE, 574 in Siemens, 767 images in Philips). For Scan2Scan, a image translation model was trained on each combination of source vendor to target vendor. For example, some of the image translation models trained were GE-to-GE (538 to 341 images), GE-Siemens (538 to 574 images), and GE-to-Philips (538 to 767 images). A summary of all the combinations can be found in Appendix under Table \ref{table:exp}

\begin{table}[!h]
\centering

\begin{tabular}{c | c | c | c | c }
\hline
\hline
\multicolumn{5}{c}{Source: MICCAI} \\
\hline
\hline
Centers & Scanner Vendors & Total & No. Labels &  No. Images \\
\hline
\hline
Amsterdam & GE & 20 & 20  & 538 \\
\hline
Singapore & Siemens & 20 & 20 & 554 \\
\hline
Utrecht & Philips & 20 & 20 & 603 \\
\hline
\hline
\multicolumn{5}{c}{Target: CAIN} \\
\hline
\hline
176, 202 & GE & 10 & 10 & 341\\
\hline
190, 1180, 1203 & Siemens & 18 & 10 & 574\\
\hline
198, 320 & Philips & 26 & 10 & 767\\
\hline

\end{tabular}
\caption{Dataset Summary}
\label{table:data}
\end{table}

\subsection{Architecture}

In Figure \ref{fig:training_setup}, S denotes the source dataset with annotations and T denotes the target dataset without annotations. The generators used consists of three convolution blocks with a stride of 2 and three ResNet blocks, as defined by \cite{zhu_unpaired_2020} and \cite{he_deep_2015}. Modifications to number of ResNet blocks were to reduce the parameter space to better suit the smaller number of images. The segmentation model is a modification of the U-Net \citep{ronneberger_u-net_2015}, consisting of five ResNet blocks \citep{he_deep_2015} in the encoder path, and mirrored with 2D transposed layers in the decoding path, with a single ResNet block as the bottleneck. 

\subsubsection{Loss Functions}
For training CycleGANs, the loss function consists of two main components. First is the adversarial loss defined as the following:

\begin{equation}
    \centering
    \mathcal{L}_{GAN}(G,D,A,B) = \mathop{{}\mathbb{E}}_{y\sim B}[{log D(y)}] + \mathop{{}\mathbb{E}}_{x\sim A}[1 - {log D(G(x))}]
    \label{eq:ad}
\end{equation}

Where, the generator attempts to generate images that approximates the real images, and the discriminator attempts to differentiate between generated and real. Second is the cycle consistency loss defined as the following: 

\begin{equation}
    \centering
    \mathcal{L}_{Cycle}(G_{A2B}, G_{B2A},A) = \mathop{{}\mathbb{E}}_{x\sim A}|{G_{B2A}(G_{A2B}(x)) - x}|_1
    \label{eq:cycle}
\end{equation}

The cycle consistency loss is a reconstruction loss following the translation through paired generators. The total loss function is defined as the following for a paired generator and discriminator: 

\begin{dmath}
    \centering
    \mathcal{L}_{Total}(G_{S2T}, G_{T2S},S,T) = \lambda_1 \mathcal{L}_{GAN}(G_{S2T}, G_{T2S},S,T) + \lambda_2 \mathcal{L}_{GAN}(G_{T2S}, G_{S2T},T,S) + \lambda_3 \mathcal{L}_{Cycle}(G_{S2T}, G_{T2S},S) + \lambda_4 \mathcal{L}_{Cycle}(G_{T2S}, G_{S2T},T)
    \label{eq:gan_total}
\end{dmath}

For all modes, we utilize the default parameters as defined in original CycleGAN \citep{zhu_unpaired_2020} for the lambda weighting in equation \ref{eq:gan_total}. For the segmentation models, we utilize generalized Dice loss \citep{sudre_generalised_2017} due to the relative class imbalance of white matter lesions in the images. 

\begin{dmath}
    \centering
    \mathcal{L}_{seg} = 1 - 2 \frac{\sum_{1=1}^2 w_l \sum_n r_{ln}p_{ln}}{\sum_{1=1}^2 w_l \sum_n r_{ln} + p_{ln}}
    \label{eq:seg_loss}
\end{dmath}

Where, r represents the real annotations and p represents the predictions. The weighting coefficient, $w_l = \frac{1}{(\sum_{n=1}^N r_{ln})^2}$, is used to correct the contribution of each class by the inverse of the volume. 

\subsubsection{Training}

For each of the image translation modes, we define the source and target distributions of the data by the different modes (Image2Image, Scan2Scan, and Label2Image). A summary of the different configurations can be found in Appendix under Table \ref{table:exp}. For each experiment, we train for 50 epochs with the last 25 epochs applying learning rate decay to 0 \citep{you_how_2019}, with an initial learning rate of 0.0002. For increasing stability, we train our discriminators with memory of eight images \citep{shrivastava_learning_2017}. The size of the batches was 4. As mentioned above, we attempt to further explore the application of Label2Image translation by attempting to utilize synthetically generated labels. To accomplish this, we train a DCGAN \citep{radford_unsupervised_2016} model using the default parameters to model the label images. Once trained, we sample 5000 images from the generator and use it as a source domain, and repeat the above experiment. 

Once the image translation models are trained, we conduct two sets of experiments for training segmentation model. The first set is creating a synthetic dataset per mode for each target distribution. For example, for Scan2Scan models that were trained with GE volumes as the target dataset, the synthetic data generated was pooled together to create a synthetic dataset. The second set consists creating multi-center synthetic dataset per mode by pooling the generated data together. For example, for the Label2Image models, the synthetic data from all the different target distributions were pooled together to make a synthetic multi-center dataset. 

The segmentation models were trained in a supervised fashion on the synthetic data with the annotations from the source domain, as shown in Figure \ref{fig:training_setup}. Only the lesions annotations were used in training. For each model, we train on 50 epochs with a batch size of 8, and utilizing Adam optimizer \citep{kingma_adam_2017} ($\beta_1 = 0.5$, and $\beta_2 = 0.999$) with an initial learning rate of 0.001. 

We repeat the segmentation training configuration to establish our baseline performance, in which we seek define a lower and upper bound. The lower bound is defined as performance of the segmentation model trained on only the source domain and tested on the target domain. This is to approximate the effect of the domain shift across different distributions. The upper bound seeks to define the performance of a segmentation model trained and tested on the target distribution data. In other words, the upper bound seeks to observe a model that is potentially over-fitted to the training data. 

\subsection{Evaluation Metrics}

To evaluate segmentation performance we utilize six evaluation metrics to compare the WML prediction masks to the ground truths. Five of the metrics are derived from standardized metrics purposed by the MICCAI 2017 WMH Segmentation Grand Challenge; Dice Similarity Coefficient (DSC), Hausdorff distance (HD), Average Volume Difference (AVD), Lesion Recall (L-Recall), and Lesion F1 (L-F1) \citep{kuijf_standardized_2019}. We also include False Positive Rate (FPR) as a percentage, since there is a significant imbalance of positive classes (lesion) and negative classes (foreground). 

One of the standard metrics to measure the quality of generated images is to utilize Fréchet inception distance (FID), which compares the Wasserstein-2 distance between two Guassians of the feature representation defined by InceptionV3 pretrained on ImageNet \citep{heusel_gans_2018}. It has been found lower scores were tended to correlate to human agreement \citep{lucic_are_2018}. The formulation of FID score is given by the following: 

\begin{equation}
    \centering
    FID = |m - m_{w}|^2 + Tr(C + C_w - 2(CC_w)^{\frac{1}{2}})
    \label{eq:fid}
\end{equation}

Where, the features are sampled from the coding layers in the InceptionV3 network, and the respective mean and co-variance are calculated for the generated images, $(m, C)$ and real world images $(m_w, C_w)$. FID score is typically calculated on sufficiently large data sets, as the estimation of the co-variance matrix requires sufficient number of samples relative to the vector length of the feature vector (2048 for original implementation). As we do not have sufficient number of samples, in this paper we explore sampling the feature vector at different depths of the network. Inspired by FID score implementation by \cite{seitzer_pytorch-fid_2020} we sample at; first max pooling features (64), second max pooling features (192), pre-auxiliary classifier (768), and final averaging pooling features (2048). While FID score is calculated using a model pre-trained on natural images may raise concerns of applicability to MR images, the objective of utilizing the metric is to have an objective feature representation to compare between the different models. 

\section{Results}

In the following, the results of experiments are presented. We examine visually the quality of the augmented data, the segmentation performance of the model trained on the augmented data, and lastly a quantitative comparison using FID score. 

\subsection{Augmented Data}

\begin{figure}[htbp]
\centering
\includegraphics[width=\linewidth]{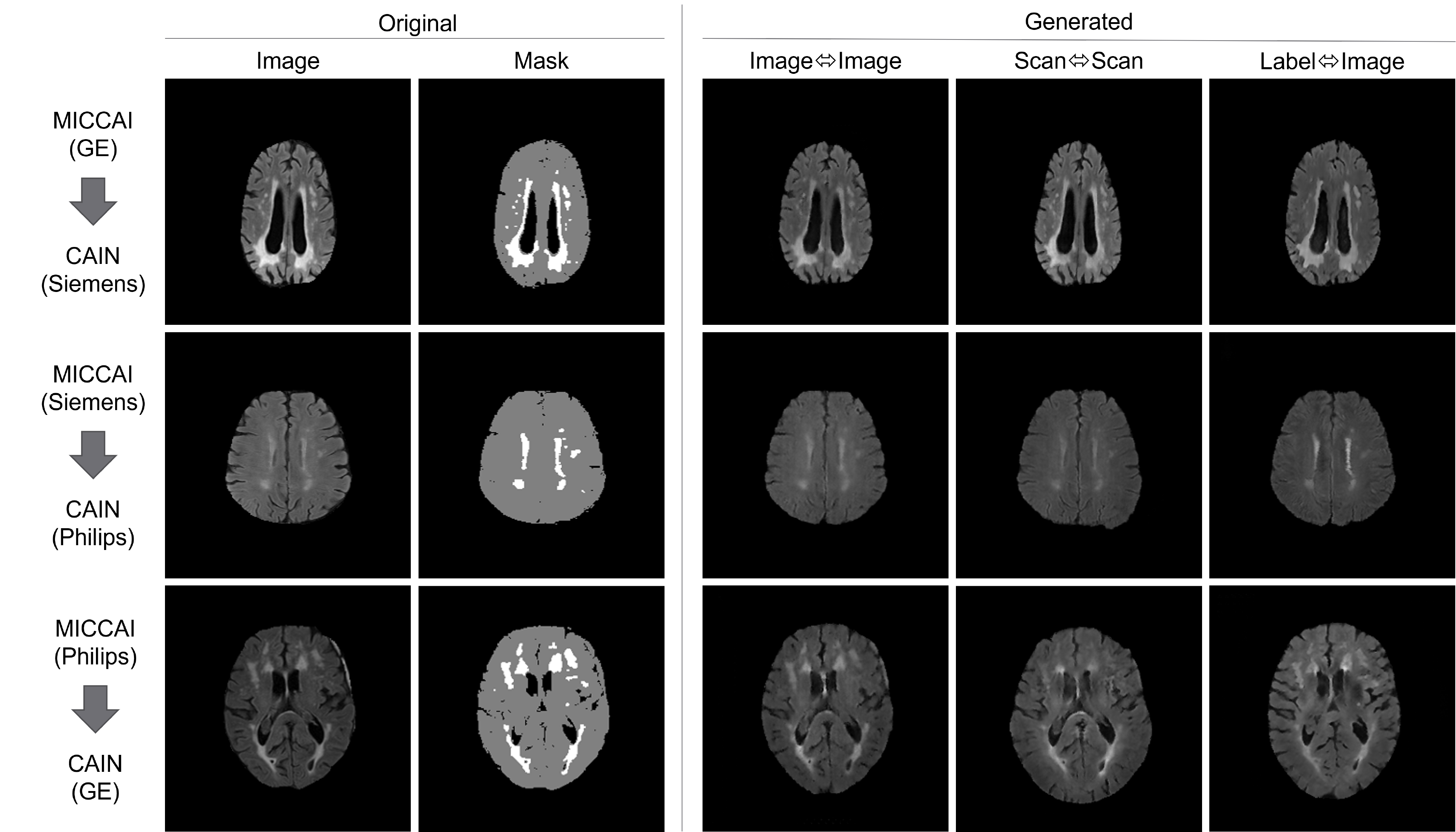}
\caption{Example of Augmentation from Multiple Priors}
\label{fig:cross_sample}
\end{figure}

Figure \ref{fig:cross_sample} shows a sample of augmentation across scanner domains. On the left, the original images are sampled from each of the centers in the source domain (MICCAI), displaying the respective MR image and mask. For the segmentation mask, white indicates WML and gray indicates brain. The augmented images on the right are the results of each translation, after augmentation they are reshaped back into the original dimension of the source image. 

\begin{figure}[htbp]
\centering
\includegraphics[width=\linewidth]{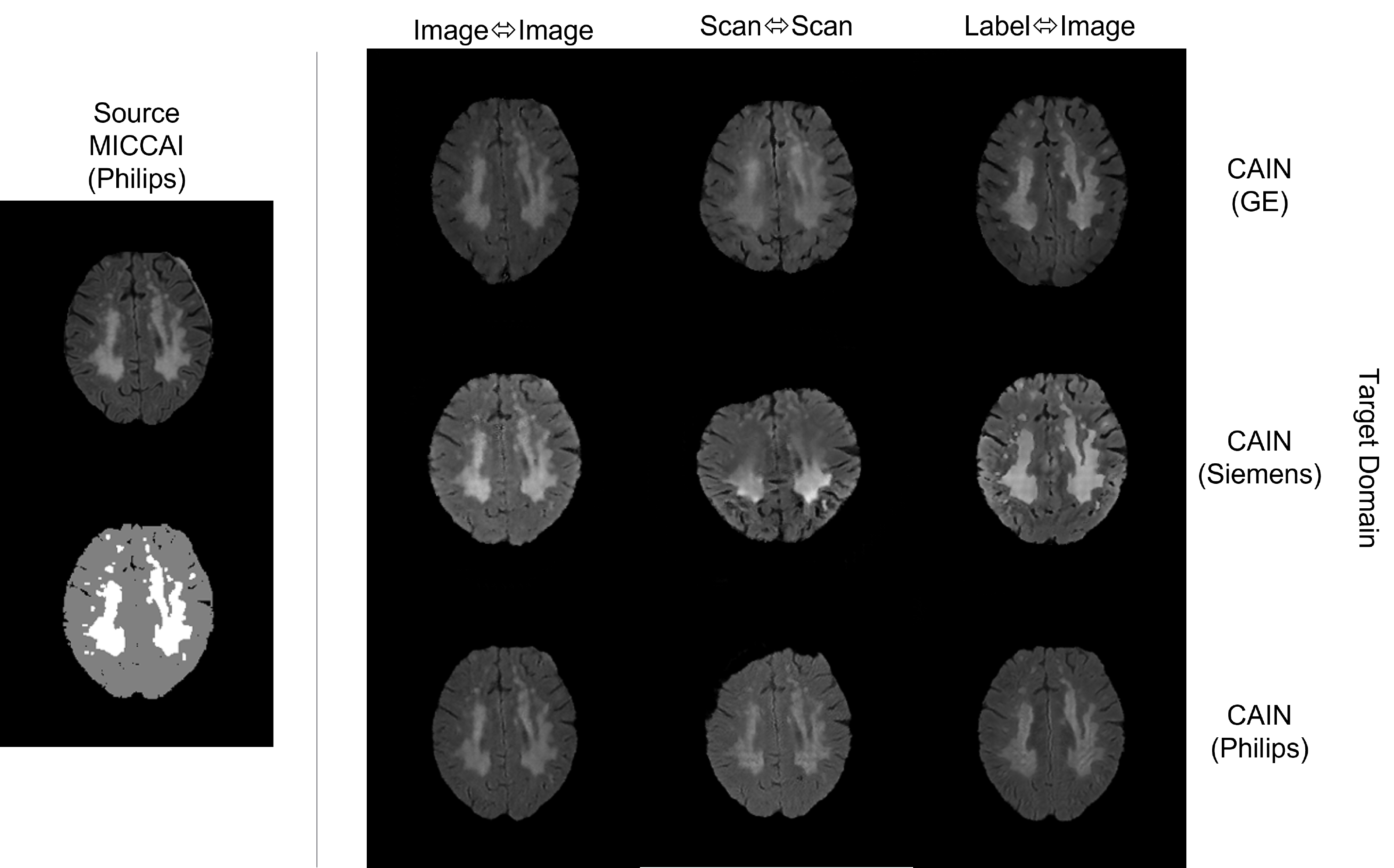}
\caption{Example of Augmentation from Single Prior}
\label{fig:single_prior}
\end{figure}

Warping is present for both Image2Image and Scan2Scan modes. For example, in Figure \ref{fig:cross_sample}, for translation between GE and Siemens, the resultant images for the two modes yields warping of the brain region when compared to original image. This is likely an artifact from augmenting from an image that has a significantly different aspect ratio.  The example shown in Figure \ref{fig:single_prior}, shows less of this particular warping due to the source image being closer to 1:1 aspect ratio. For Scan2Scan, there appears to be cases in where the labels are not maintained through translation, as shown in Philips to Siemens translation (center image) in Figure \ref{fig:single_prior}. The Label2Image translation appears to be the most consistent in terms of overall shape and label consistency. Figure \ref{fig:cross_sample} and \ref{fig:single_prior} both show relatively realistic appearing images. However, some anatomically incorrect gyri appear in the example in the Philips to GE translation (top right) in Figure \ref{fig:single_prior}.

\subsubsection{Augmentation from Synthetic Labels}

\begin{figure}[htbp]
\centering
\includegraphics[width=\linewidth]{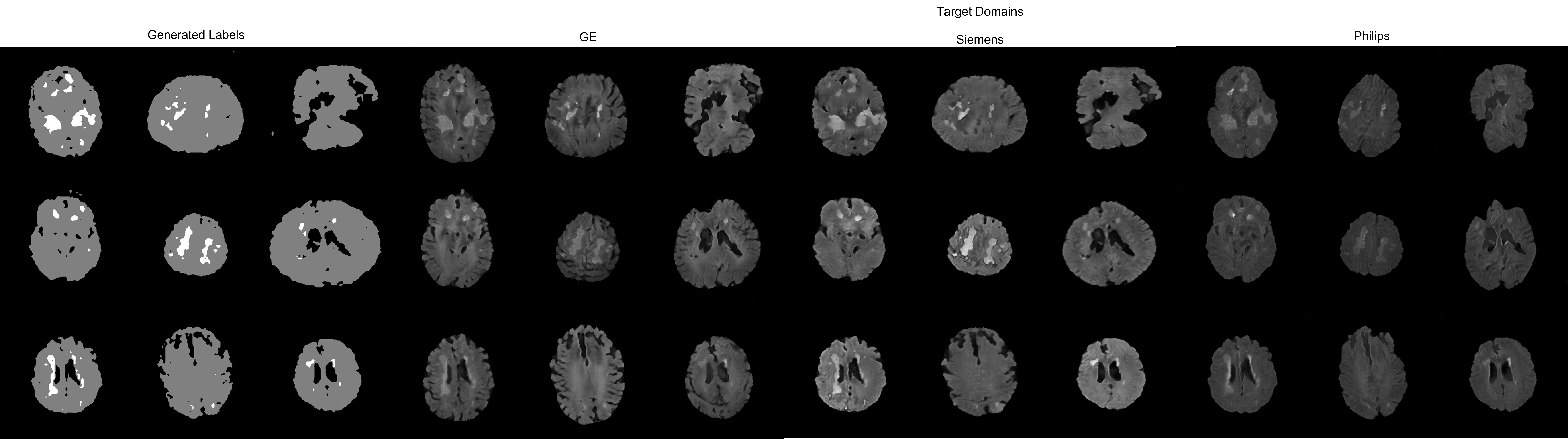}
\caption{Synthetic Labels to Target Domains}
\label{fig:syn_label_samples}
\end{figure}

Extending the investigation of utilizing labels as priors for translation, Figure \ref{fig:syn_label_samples} displays the results of the synthetic labels generated on the left and the resultant translation to each domain on the right. As with examples shown in Figure \ref{fig:cross_sample} and \ref{fig:single_prior}, the overall shape and label consistency is maintained relative to original input, performing similarly to Label2Image in the above section. Overall brain images appear to be easily discernible as generated images. Some portions of the generation appears less textured than what is typically expected in real images as well, leading to suggestion that the model may be overgeneralizing. 

\subsection{Segmentation Performance}

\subsubsection{Training on Target Distribution}
\begin{figure}[htbp]
\centering
\includegraphics[width=\linewidth]{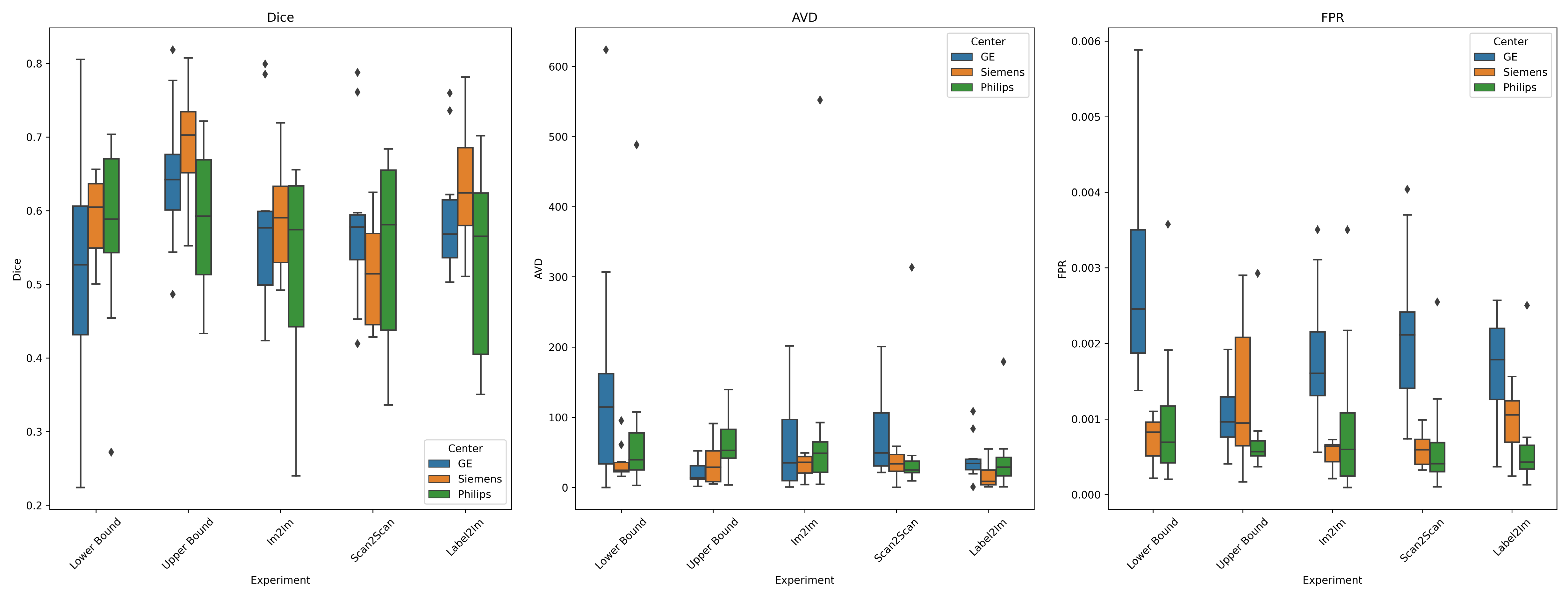}
\caption{Segmentation Performance for Single Domain}
\label{fig:target_training}
\end{figure}

In this section, the results for the set of experiments where segmentation models were trained on synthetic data derived from a single scanner in the target distribution will be analyzed in the context of lesion segmentation. For each augmentation mode, the segmentation performance for each scanner vendor is shown in Figure \ref{fig:target_training}. Observing the overall trend in Figure \ref{fig:target_training}, across all the models, there appears to be significant improvement when comparing performance of segmentation on GE volumes to the lower bound. There appears to be slight improvement in segmentation on Siemens volumes for Image2Image and Label2Image models, with notable performance of Label2Image performance being similar to upper bound performance. Dice similarity performance for Philips volumes appears to have worsened across all models, however AVD and FPR still show positive results with low scores for Philips. In Table \ref{table:target_table} the performance of the individual models are summarized for the respective metrics. Also within Table \ref{table:target_table}, we show segmentation results where we have taken the liberty to explore the possibility of utilizing CycleGANs as a form of unsupervised segmentation model for the Label2Image models. Since, for the training of Label2Image, there exists a Label2Image model that is paired with it, which acts in a similar fashion as training segmentation. However, the segmentation provided in this unsupervised manner was not viable, as the average Dice coefficients scores were 0.44, 0.37, and 0.32 for the respective scanners (GE, Siemens, Philips). 

\subsubsection{Training on Mixed Distribution}

\begin{figure}[htbp]
\centering
\includegraphics[width=\linewidth]{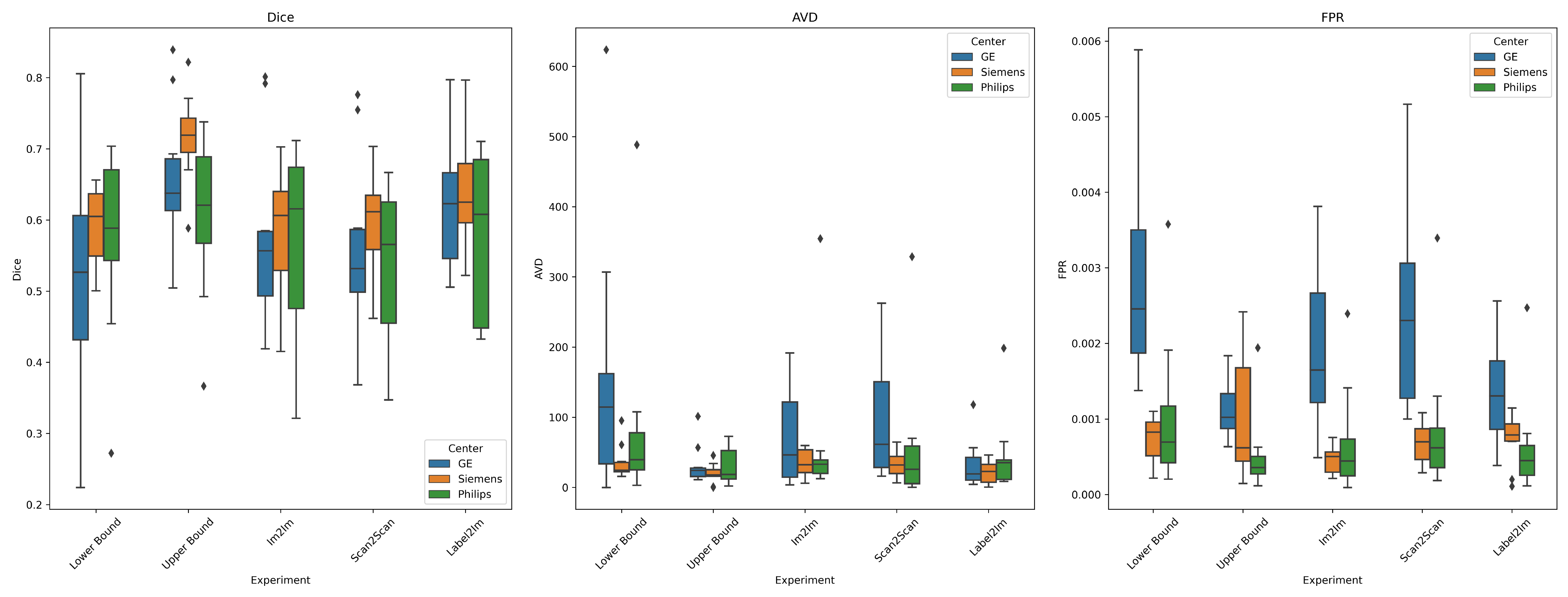}
\caption{Segmentation Performance for Mixed Domain}
\label{fig:mixed_training}
\end{figure}

\begin{figure}[htbp]
\centering
\includegraphics[width=0.5\linewidth]{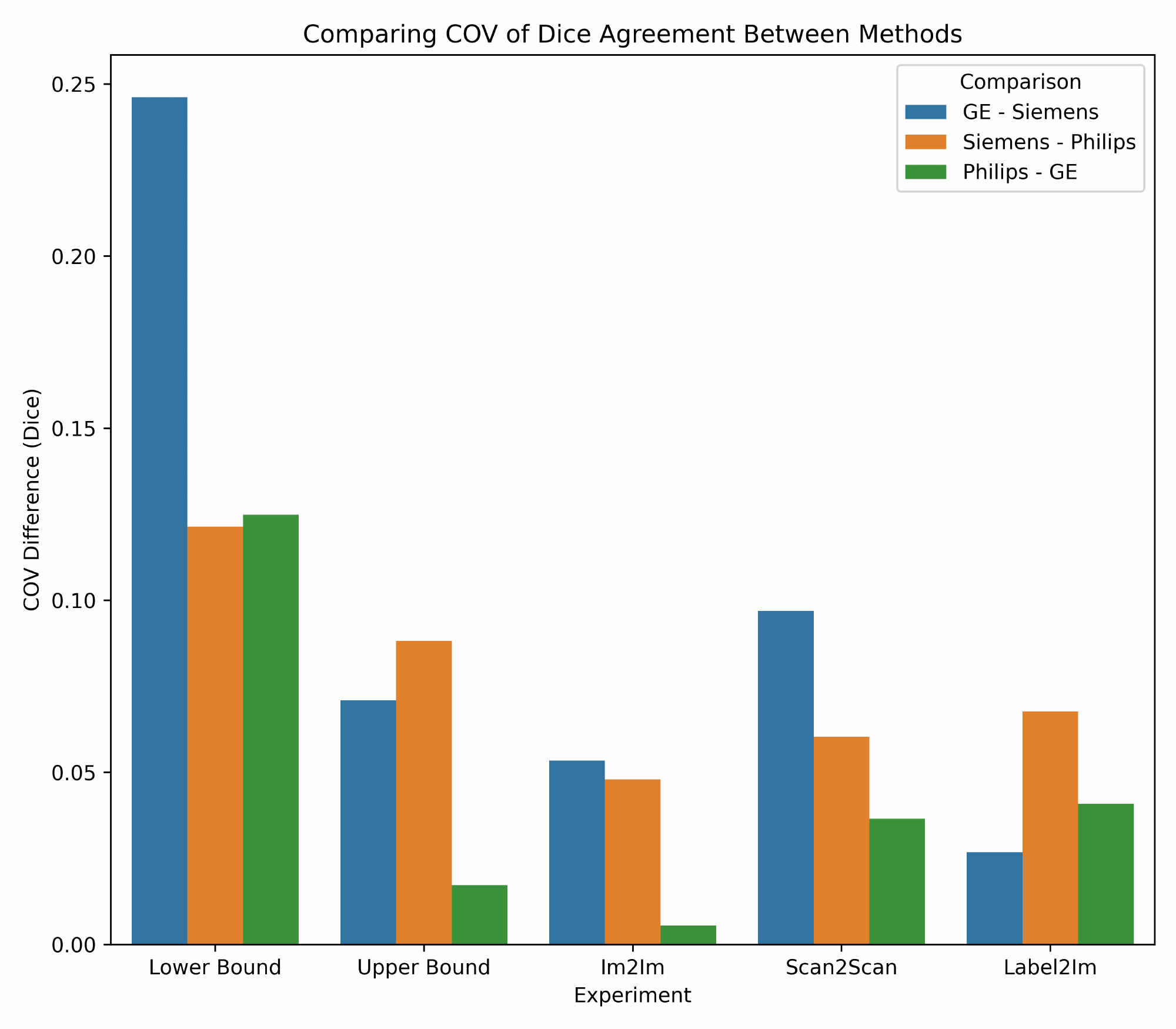}
\caption{Comparison of COV between centers for each method}
\label{fig:cov_diff}
\end{figure}

In the following, the lesion segmentation performance will be analysed for the set of experiments where segmentation models were trained on synthetic multi-center data, for which synthetic data generated by each augmentation mode were pooled together. For each augmentation mode, the segmentation performance for each scanner vendor is shown in Figure \ref{fig:mixed_training}. It was expected that there would be improvement to the upper bound performance due to larger dataset. Comparing Figure \ref{fig:target_training} and \ref{fig:mixed_training}, for the Image2Image and Scan2Scan models, there appears to be no significant improvement on GE and Siemens domains, however performance on Philips did improve slightly. Notably, the performance of the models trained on label-to-image models show significant improvement on GE and Philips domains, with 0.04 and 0.05 improvement respectively on DSC, with a minor improvement on Siemens domain, with an improvement of 0.01. In Figure \ref{fig:cov_diff}, we present the relative consistency between scanner vendors for each mode as a function of coefficient of variation (COV), which is defined as the ratio of the standard deviation to the mean \citep{everitt_cambridge_2002}. We observe that for each mode we investigated, there is  significant decrease in overall COV relative to the COV of the lower bound, indicating an overall improvement to consistency. 

The results above imply that the Label2Image is the optimal mode for image translation in context of generating synthetic samples for segmentation training. Exploring this implication further, the experiments were repeated for the data generated in Figure \ref{fig:syn_label_samples}. The results presented in Appendix under Table \ref{table:mixed_table} show that the these series of experiments did not yield good segmentation performance, with average Dice coefficients of 0.40, 0.40, and 0.32 for GE, Siemens, and Philips respectively compared to Dice coefficient averages by Label2Image of 0.63, 0.64, and 0.58. 

\subsection{FID and Visualization}

\begin{figure}[htbp]
\centering
\includegraphics[width=\linewidth]{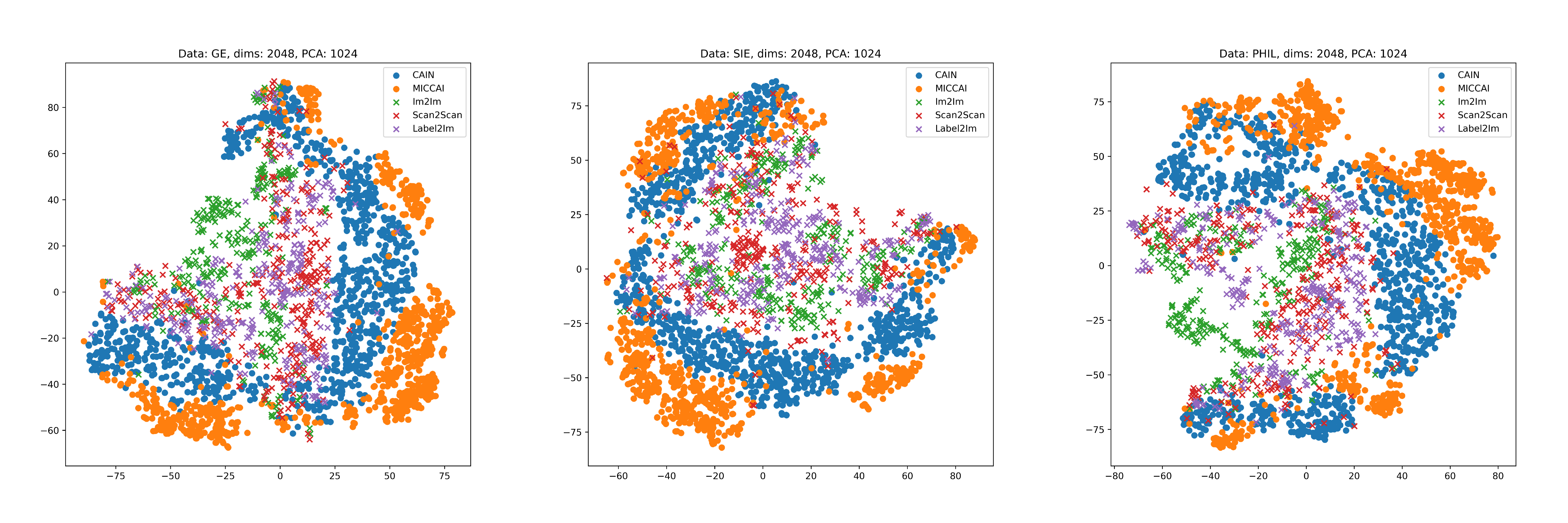}
\caption{T-SNE Visualization}
\label{fig:tsne}
\end{figure}

Analysis in the following attempts to quantify the visual quality between the augmentation modes as compared to the real images from the target (CAIN) distribution. Table \ref{table:fid} shows the FID scores relative to the target (CAIN) distribution of each of the modes and the source (MICCAI) dataset  at different depths of the InceptionV3 network. The exploration of different depths was to observe the comparison of the features at different levels of complexity (with lower layers consisting of low complexity features, to more complex features at later layers). We observe the scores of Image2Image mode appears to have the highest FID score for the deeper layers, even when comparing to scores generated by MICCAI. Scan2Scan images appear to have the lowest FID scores, possibly indicating the images being closer to the target distribution. 

Figure \ref{fig:tsne}, shows a visualization of the features sampled at the final averaging pooling features using T-SNE \citep{van_der_maaten_visualizing_2008}. Since the images are MR images, the expected natural image features are likely to be redundant and not applicable, thus we first utilize PCA to reduce the dimensions from 2048 to 1024 \citep{jolliffe_principal_2016}. For each of the domains, the generated images tend to be cluster around each other, while being closer to the target (CAIN) distribution, while being further from source dataset (MICCAI). 

\begin{table}[htbp]
\centering
\begin{tabular}{l|l|l|l|l}
\hline
\hline
Method    & FID: 64 & FID: 128 & FID: 768 & FID: 2048 \\
\hline
\hline
Image2Image     & 0.017   & 0.299    & 0.161    & 65.129    \\\hline
Scan2scan & 0.012   & 0.164    & 0.091    & 36.493    \\\hline
Label2Image  & 0.014   & 0.154    & 0.097    & 39.393    \\\hline
MICCAI    & 0.318   & 1.289    & 0.146    & 46.975    \\\hline
\end{tabular}
\caption{Modified FID score along different vector lengths}
\label{table:fid}
\end{table}

\section{Discussion}

It was demonstrated that data points representative of a target dataset can be generated through augmenting a source dataset through image translation. By augmenting the data and reusing the annotation labels, we found that it is possible to approach segmentation performance defined by the upper bound model (trained on the target distribution directly). Our investigation found that synthetic data generated from label images yielded results comparable to segmentation models trained directly on the target dataset. In this section, we discuss further on the multi-center effect, comparison between label versus real images as priors, and the importance of realism in training. 

\subsection{The Multi-Center Effect}

It was expected that multi-center data would have increased variability due to the multi-center effect. We observe this in Figures \ref{fig:target_training} and \ref{fig:mixed_training}, when observing the upper bound model (which was trained directly on the target dataset) having higher variability in segmentation performance in terms of Dice coefficient agreement. The higher variability can be understood as the problem of concept drift \citep{zenisek_machine_2019}, in where systems become less accurate over time as more data is introduced. However, this effect is not well understood in context of image synthesis. We attempt to observe this effect by changing the variability of the source datasets for the case of Image2Image and Scan2Scan. Presented in Figure \ref{fig:cross_sample} and \ref{fig:single_prior}, the generated samples of Image2Image and Scan2Scan can be observed to create relatively realistic images that are difficult to discern visually which are better. Observing the segmentation performance of models trained on this data in Figure \ref{fig:target_training} and \ref{fig:mixed_training} show similar Dice agreement between Image2Image and Scan2Scan models. Therefore, it may be inconclusive whether the multi-center effect is present in image synthesis for the case presented in this paper. However, the relative amount of data used in training may suggest overfitting in generative models. In study by \cite{karras_training_2020}, generative models were monitored by their FID score to a test set and varied by the amount of data used in training. It was found that models with less training data tended to overfit, as the FID scores diverged at earlier points than models trained with more data. From this we can infer that the generated images tended to contain features more specific to the training set, and in which we observe this in Table \ref{table:fid}, where Scan2Scan models show significantly lower FID scores than Image2Image models. 

\subsection{Label versus Images as Priors}

The cycle consistency introduced in CycleGAN \citep{zhu_unpaired_2020} regularizes the training of the model by enforcing the translation $G_{B2A}(G_{A2B}(x)) \approx x$. What this infers, is that information and features in image $x$ is encoded by $G_{A2B}(.)$, such that it can be decoded by the paired network, $G_{B2A}(.)$. For both Image2Image and Scan2Scan modes, as their image priors were real images, the expected information maintained through translation would cause the generated images to be similar. We can observe the relative similarity to the prior images for Image2Image and Scan2Scan modes in Figure \ref{fig:cross_sample} and \ref{fig:single_prior}. However, despite the generated samples being more similar to real images, Figure \ref{fig:target_training} and \ref{fig:mixed_training} show that segmentation models trained on Label2Image performed better in every segmentation metric and approximates better to the performance of model trained directly on target data. One suggestion of why Label2Image data perform better is that the label image priors are less complex and contain less information than real priors. Lower complexity means that less features have to be maintained in the translation of $G_{A2B}(x) = x^'$, and features in synthetic image $x^'$ primarily consist of features learned of the target distribution. Another reason may be due to the function of the decoder network to maintain label consistency. Since the forward path, $G_{A2B}(x) = x^'$, creates a synthetic MR image, the decoder path, $G_{A2B}(x^') = x$, translates the synthetic image back to label domain. In effect, the decoder path functions similarly to a segmentation model. Since the decoder path functions similarly as a segmentation model, this opens up the consideration of unsupervised learning for segmentation. However, we evaluate the decoders respective of each model for Label2Image in Appendix under Table \ref{table:target_table} as a function of segmentation performance and was found that unsatisfactory results, with unsupervised performance scoring lower than our segmentation model trained on out-of-domain data. 

\subsection{Realism in Training}

Relative to images generated by Image2Image and Scan2Scan modes, Label2Image data appears to be less realistic, with overgeneralized textures and implausible structures appearing in generated data. This is likely due to the label image priors containing less details and are therefore less restrictive in translation. Despite this, the segmentation models trained on Label2Image data had the highest performance. It may be suggested that by training on overgeneralized images, the segmentation model is better trained to learn features of the target distribution. To explain further, this could be due to the operations of convolution layers in neural networks, as the overall receptive field is activated when a learned feature is "detected", the relative location within the receptive field may not matter. To observe the limits of this, we experimented with generating synthetic labels and using them a priors for Label2Image translation. Figure \ref{fig:syn_label_samples} shows a sample of the synthetic labels translated to each respective target distribution. In these samples, the generated images are less realistic and contain more implausible strictures. The results of the training of these generated data can be found in Appendix under Table \ref{table:mixed_table}. The segmentation performance of models trained with this data was lower than the model trained on out-of-domain data, suggesting there may be a minimum requirement to realism that needs to be met. 

\section{Conclusion}

In this paper, it was demonstrated that CycleGAN \citep{zhu_unpaired_2020} can be implemented to create synthetic data points representative of a target dataset through image translation, and said data points can be used to train segmentation models. Specifically, we considered the application for augmentation between MR dataset, and investigate the particular configurations of the distributions between source and target. The performance of these segmentation models have shown to approach performances similar to segmentation models trained directly on the target dataset. It was observed that the optimal configuration is to utilize label images as a prior to create synthetic data points, despite the generated images being less realistic and overgeneralized. Future work intends to explore the relationship between realism in synthetic data and applicability in augmenting small datasets. Potential applications could be found in topics such as active learning, adversarial attacks, and pseudo-labeling. Within the clinical setting, training data representative of a particular vendor or center could potentially be readily generated from a few samples. In turn, this could reduce the overhead required to train robust algorithms and make the overall process of integration more scalable. 


\acks{We thank NSERC Discovery Grant Program for funding this research.}

%
\ethics{The work follows appropriate ethical standards in conducting research and writing the manuscript, following all applicable laws and regulations regarding treatment of animals or human subjects.}

\coi{We declare we don't have conflicts of interest.}

\newpage

\appendix 
\section*{Appendix A.}

\subsection{Segmentation Model}

\begin{figure}[!htbp]
\centering
\includegraphics[width=0.75\linewidth]{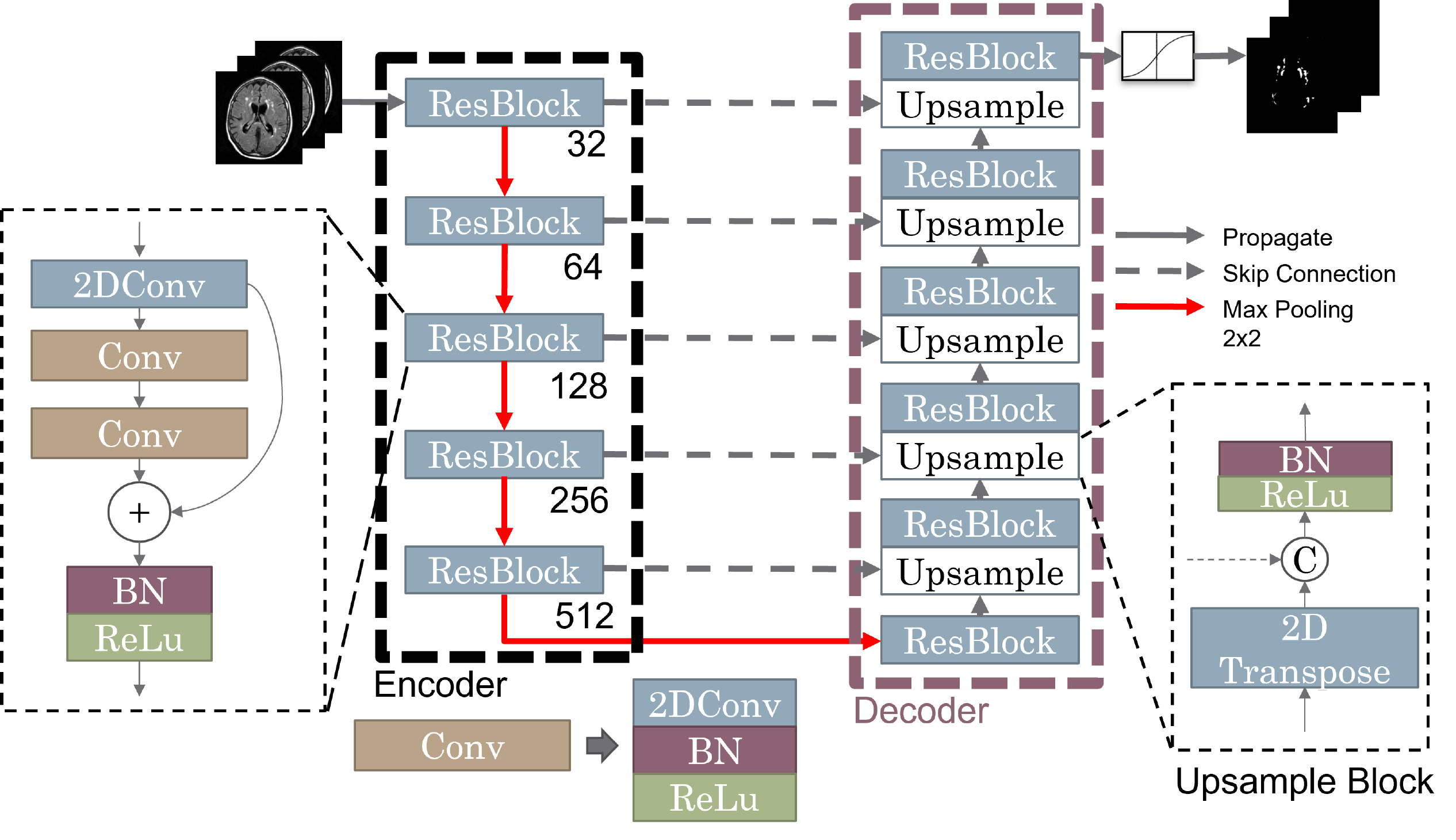}
\caption{Segmentation Model}
\label{fig:seg}
\end{figure}

\newpage

\subsection{Data Table for Each Experiment}

\begin{table}[!htbp]
\centering
\begin{tabular}{c|c|c|c|c}
\hline \hline
Mode        & Source          & No. Images & Target        & No. Images \\ \hline \hline
Image2Image & MICCAI          & 1695       & CAIN: GE      & 341        \\ \hline
Image2Image & MICCAI          & 1695       & CAIN: Simens  & 574        \\ \hline
Image2Image & MICCAI          & 1695       & CAIN: Philips & 767        \\ \hline \hline
Scan2Scan & MICCAI: GE      & 538        & CAIN: GE      & 341        \\ \hline
Scan2Scan & MICCAI: GE      & 538        & CAIN: Siemens & 574        \\ \hline
Scan2Scan & MICCAI: GE      & 538        & CAIN: Philips & 767        \\ \hline
Scan2Scan & MICCAI: Siemens & 554        & CAIN: GE      & 341        \\ \hline
Scan2Scan & MICCAI: Siemens & 554        & CAIN: Simens  & 574        \\ \hline
Scan2Scan & MICCAI: Siemens & 554        & CAIN: Philips & 767        \\ \hline
Scan2Scan & MICCAI: Philips & 603        & CAIN: GE      & 341        \\ \hline
Scan2Scan & MICCAI: Philips & 603        & CAIN: Simens  & 574        \\ \hline
Scan2Scan & MICCAI: Philips & 603        & CAIN: Philips & 767        \\ \hline \hline
Label2Image & MICCAI: Labels  & 1695       & CAIN: GE      & 341        \\ \hline
Label2Image & MICCAI: Labels  & 1695       & CAIN: Simens  & 574        \\ \hline
Label2Image & MICCAI: Labels  & 1695       & CAIN: Philips & 767        \\ \hline \hline
DCGAN   & N/A      & N/A       & MICCAI: Labels      & 1695        \\ \hline \hline
Syn2Image   & Syn Labels      & 5000       & CAIN: GE      & 341        \\ \hline 
Syn2Image   & Syn Labels      & 5000       & CAIN: Simens  & 574        \\ \hline
Syn2Image   & Syn Labels      & 5000       & CAIN: Philips & 767        \\ \hline
\end{tabular}
\caption{Summary of Data Splits for GAN Training}
\label{table:exp}
\end{table}

\begin{table}[]
    \centering
    \begin{tabular}{c|c|c}
        \hline \hline
        \multicolumn{3}{c}{Single Target Distribution} \\
        \hline \hline
        Mode        & Sources          & No. Images \\ \hline \hline
        Image2Image & MICCAI $\rightarrow$ CAIN: GE (1695) & 1695 \\ \hline
        Image2Image & MICCAI $\rightarrow$ CAIN: Siemens (1695) & 1695 \\ \hline
        Image2Image & MICCAI $\rightarrow$ CAIN: Philips (1695) & 1695 \\ \hline 
        \hline
        Scan2Scan & \makecell{MICCAI: GE $\rightarrow$ CAIN: GE (538) + \\ 
                            MICCAI: Siemens $\rightarrow$ CAIN: GE (554) + \\ 
                            MICCAI: Philips $\rightarrow$ CAIN: GE (603)} & 1695 \\ \hline
        Scan2Scan & \makecell{MICCAI: GE $\rightarrow$ CAIN: Siemens (538) + \\ 
                            MICCAI: Siemens $\rightarrow$ CAIN: Siemens (554) + \\ 
                            MICCAI: Philips $\rightarrow$ CAIN: Siemens (603)} & 1695 \\ \hline
        Scan2Scan & \makecell{MICCAI: GE $\rightarrow$ CAIN: Philips (538) + \\ 
                            MICCAI: Siemens $\rightarrow$ CAIN: Philips (554) + \\ 
                            MICCAI: Philips $\rightarrow$ CAIN: Philips (603)} & 1695 \\ \hline  
        \hline
        Label2Image & MICCAI: Labels $\rightarrow$ CAIN: GE (1695) & 1695 \\ \hline
        Label2Image & MICCAI: Labels $\rightarrow$ CAIN: Siemens (1695) & 1695 \\ \hline
        Label2Image & MICCAI: Labels $\rightarrow$ CAIN: Philips (1695) & 1695 \\ \hline 
        \hline
        Syn2Image & Syn Labels $\rightarrow$ CAIN: GE (5000) & 5000 \\ \hline
        Syn2Image & Syn Labels $\rightarrow$ CAIN: Siemens (5000) & 5000 \\ \hline
        Syn2Image & Syn Labels $\rightarrow$ CAIN: Philips (5000) & 5000 \\  
        \hline \hline
        \multicolumn{3}{c}{Mixed Target Distribution} \\
        \hline \hline
        Image2Image & \makecell{MICCAI $\rightarrow$ CAIN: GE (1695) + \\ 
                            MICCAI $\rightarrow$ CAIN: Siemens (1695) + \\ 
                            MICCAI $\rightarrow$ CAIN: Philips (1695)} & 5085 \\ \hline 
        Scan2Scan & \makecell{MICCAI: GE $\rightarrow$ CAIN: GE (538) + \\ 
                            MICCAI: Siemens $\rightarrow$ CAIN: GE (554) + \\ 
                            MICCAI: Philips $\rightarrow$ CAIN: GE (603) + \\
                            MICCAI: GE $\rightarrow$ CAIN: Siemens (538) + \\ 
                            MICCAI: Siemens $\rightarrow$ CAIN: Siemens (554) + \\ 
                            MICCAI: Philips $\rightarrow$ CAIN: Siemens (603) + \\
                            MICCAI: GE $\rightarrow$ CAIN: Philips (538) + \\ 
                            MICCAI: Siemens $\rightarrow$ CAIN: Philips (554) + \\ 
                            MICCAI: Philips $\rightarrow$ CAIN: Philips (603)} & 5085 \\ \hline 
        Label2Image & \makecell{MICCAI: Labels $\rightarrow$ CAIN: GE (1695) + \\ 
                            MICCAI: Labels $\rightarrow$ CAIN: Siemens (1695) + \\ 
                            MICCAI: Labels $\rightarrow$ CAIN: Philips (1695)} & 5085 \\ \hline  
        Syn2Image & \makecell{Syn Labels $\rightarrow$ CAIN: GE (5000) + \\ 
                            Syn Labels $\rightarrow$ CAIN: Siemens (5000) + \\ 
                            Syn Labels $\rightarrow$ CAIN: Philips (5000)} & 15000 \\ \hline 
    \end{tabular}
    \caption{Summary of Synthetic Datasets for Segmentation Training}
    \label{table:syn_data_pools}
\end{table}

\newpage
\subsection{Segmentation Performance}

\subsubsection{Training on Single Distribution}

\begin{table}[!ht]
\centering
\begin{tabular}{l|l|l|l|l|l|l}
\hline
\hline
Method    & Dice      & HD          & AVD           & L-Recall  & L-F1      & FPR (\%)  \\ \hline
\hline
\multicolumn{7}{c}{Target: GE}                                                        \\ \hline \hline
Lower     & 0.53±0.18 & 11.02±5.46  & 156.57±187.69 & 0.80±0.14 & 0.46±0.18 & 0.29±0.15 \\ \hline
Upper     & 0.65±0.10 & 7.40±35.04  & 22.09±18.44   & 0.61±0.18 & 0.67±0.12 & 0.10±0.05 \\ \hline
Image2Image     & 0.58±0.13 & 8.48±66.04  & 65.07±72.30   & 0.73±0.16 & 0.54±0.11 & 0.18±0.10 \\ \hline
Scan2scan & 0.58±0.12 & 9.14±72.25  & 73.72±56.91   & 0.78±0.13 & 0.51±0.10 & 0.22±0.11 \\ \hline
Label2Image  & 0.59±0.09 & 9.99±72.13  & 41.16±31.69   & 0.75±0.13 & 0.56±0.12 & 0.17±0.07 \\ \hline
Unsupervised & 0.44±0.12 & 13.40±4.60  & 44.93±25.82   & 0.51±0.14 & 0.24±0.13 & 0.21±0.08 \\ \hline 
\hline
\multicolumn{7}{c}{Target: Siemens}                                                   \\ \hline \hline
Lower     & 0.59±0.06 & 19.50±5.49  & 35.50±24.62   & 0.66±0.15 & 0.47±0.09 & 0.07±0.03 \\ \hline
Upper     & 0.69±0.07 & 11.93±9.75  & 32.90±28.80   & 0.61±0.15 & 0.56±0.14 & 0.13±0.09 \\ \hline
Image2Image     & 0.59±0.07 & 17.07±7.21  & 31.82±15.62   & 0.58±0.19 & 0.51±0.12 & 0.05±0.02 \\ \hline
Scan2scan & 0.51±0.07 & 14.33±4.32  & 32.24±19.23   & 0.61±0.15 & 0.44±0.11 & 0.06±0.02 \\ \hline
Label2Image  & 0.63±0.08 & 17.35±6.66  & 16.11±17.03   & 0.59±0.14 & 0.45±0.10 & 0.10±0.04 \\ \hline
Unsupervised & 0.37±0.14 & 24.21±4.30  & 37.54±17.59   & 0.29±0.12 & 0.19±0.05 & 0.08±0.04 \\ \hline
\hline
\multicolumn{7}{c}{Target: Philips}                                                   \\ \hline \hline
Lower     & 0.57±0.13 & 23.94±9.25  & 83.75±131.35  & 0.73±0.18 & 0.60±0.12 & 0.10±0.10 \\ \hline
Upper     & 0.60±0.09 & 14.44±6.30  & 62.79±42.44   & 0.75±0.18 & 0.69±0.07 & 0.08±0.07 \\ \hline
Image2Image     & 0.52±0.14 & 28.46±14.09 & 85.85±149.31  & 0.49±0.18 & 0.50±0.10 & 0.09±0.10 \\ \hline
Scan2scan & 0.55±0.13 & 24.14±16.59 & 50.11±83.72   & 0.52±0.18 & 0.57±0.10 & 0.07±0.07 \\ \hline
Label2Image  & 0.53±0.13 & 34.37±19.54 & 39.98±46.72   & 0.47±0.16 & 0.52±0.10 & 0.06±0.06 \\ \hline
Unsupervised & 0.32±0.13 & 37.37±12.24 & 40.24±30.56   & 0.31±0.17 & 0.25±0.04 & 0.06±0.08 \\ \hline
\end{tabular}
\caption{Training on Target Distribution}
\label{table:target_table}
\end{table}

\newpage
\subsubsection{Training on Mixed Distribution}

\begin{table}[!ht]
\centering
\begin{tabular}{l|l|l|l|l|l|l}
\hline
\hline
Method    & Dice      & HD          & AVD           & L-Recall  & L-F1      & FPR (\%)  \\ \hline
\hline
\multicolumn{7}{c}{Target: GE}                                                        \\ \hline \hline
Lower     & 0.53±0.18 & 11.01±5.46  & 156.57±187.69 & 0.80±0.14 & 0.46±0.18 & 0.29±0.15 \\ \hline
Upper     & 0.65±0.10 & 7.65±4.17   & 31.94±27.51   & 0.67±0.21 & 0.69±0.10 & 0.11±0.04 \\ \hline
Image2Image     & 0.58±0.13 & 9.02±4.71   & 73.91±73.19   & 0.74±0.17 & 0.53±0.10 & 0.19±0.11 \\ \hline
Scan2scan & 0.56±0.13 & 7.80±3.72   & 93.37±82.55   & 0.76±0.16 & 0.58±0.09 & 0.26±0.15 \\ \hline
Label2Image  & 0.63±0.10 & 8.07±3.37   & 33.05±34.69   & 0.72±0.17 & 0.63±0.07 & 0.14±0.07 \\ \hline
Syn2Im       & 0.40±0.14 & 16.44±6.77  & 34.20±28.59   & 0.23±0.10 & 0.33±0.11 & 0.13±0.07 \\ \hline
\hline
\multicolumn{7}{c}{Target: Siemens}                                                   \\ \hline \hline
Lower     & 0.59±0.06 & 19.50±5.49  & 35.50±24.62   & 0.66±0.15 & 0.47±0.09 & 0.07±0.03 \\ \hline
Upper     & 0.72±0.06 & 5.66±3.70   & 19.72±13.79   & 0.61±0.12 & 0.69±0.09 & 0.10±0.08 \\ \hline
Image2Image     & 0.58±0.10 & 9.63±2.02   & 35.06±20.35   & 0.61±0.14 & 0.55±0.08 & 0.05±0.02 \\ \hline
Scan2scan & 0.59±0.08 & 8.64±5.14   & 32.53±18.14   & 0.65±0.15 & 0.59±0.09 & 0.07±0.03 \\ \hline
Label2Image  & 0.64±0.08 & 15.70±7.74  & 22.17±16.13   & 0.62±0.11 & 0.48±0.10 & 0.07±0.03 \\ \hline
Syn2Im       & 0.40±0.17 & 27.13±18.74 & 23.68±26.06   & 0.18±0.07 & 0.26±0.08 & 0.12±0.06 \\ \hline
\hline
\multicolumn{7}{c}{Target: Philips}                                                   \\ \hline \hline
Lower     & 0.57±0.13 & 23.94±9.25  & 83.75±131.35  & 0.73±0.18 & 0.60±0.12 & 0.10±0.10 \\ \hline
Upper     & 0.61±0.11 & 20.15±9.28  & 29.31±25.08   & 0.49±0.16 & 0.62±0.12 & 0.05±0.05 \\ \hline
Image2Image     & 0.58±0.13 & 25.13±16.43 & 57.16±94.40   & 0.56±0.20 & 0.57±0.11 & 0.07±0.07 \\ \hline
Scan2scan & 0.54±0.10 & 18.89±9.90  & 54.48±89.86   & 0.65±0.18 & 0.62±0.11 & 0.08±0.09 \\ \hline
Label2Image  & 0.58±0.12 & 25.34±20.01 & 42.72±51.90   & 0.60±0.20 & 0.61±0.10 & 0.06±0.06 \\ \hline
Syn2Im       & 0.32±0.12 & 52.52±33.10 & 69.49±111.64  & 0.17±0.12 & 0.26±0.14 & 0.08±0.06 \\ \hline
\end{tabular}
\caption{Training on Mixed Distribution}
\label{table:mixed_table}
\end{table}

\vskip 0.2in
\newpage
\bibliography{main.bbl}

\begin{thebibliography}{51}
\providecommand{\natexlab}[1]{#1}
\providecommand{\url}[1]{\texttt{#1}}
\expandafter\ifx\csname urlstyle\endcsname\relax
  \providecommand{\doi}[1]{doi: #1}\else
  \providecommand{\doi}{doi: \begingroup \urlstyle{rm}\Url}\fi

\bibitem[Bearman et~al.(2016)Bearman, Russakovsky, Ferrari, and
  Fei-Fei]{bearman_whats_2016}
Amy Bearman, Olga Russakovsky, Vittorio Ferrari, and Li~Fei-Fei.
\newblock What's the {Point}: {Semantic} {Segmentation} with {Point}
  {Supervision}.
\newblock \emph{arXiv:1506.02106 [cs]}, July 2016.
\newblock URL \url{http://arxiv.org/abs/1506.02106}.
\newblock arXiv: 1506.02106.

\bibitem[DiGregorio et~al.(2021)DiGregorio, Arezza, Gibicar, Moody, Tyrrell,
  and Khademi]{digregorio_intracranial_2021}
Justin DiGregorio, Giordano Arezza, Adam Gibicar, Alan~R. Moody, Pascal~N.
  Tyrrell, and April Khademi.
\newblock Intracranial volume segmentation for neurodegenerative populations
  using multicentre {FLAIR} {MRI}.
\newblock \emph{Neuroimage: Reports}, 1\penalty0 (1):\penalty0 100006, March
  2021.
\newblock ISSN 26669560.
\newblock \doi{10.1016/j.ynirp.2021.100006}.
\newblock URL
  \url{https://linkinghub.elsevier.com/retrieve/pii/S2666956021000040}.

\bibitem[Dinsdale et~al.(2021)Dinsdale, Jenkinson, and
  Namburete]{dinsdale_deep_2021}
Nicola~K. Dinsdale, Mark Jenkinson, and Ana~I.L. Namburete.
\newblock Deep learning-based unlearning of dataset bias for {MRI}
  harmonisation and confound removal.
\newblock \emph{NeuroImage}, 228:\penalty0 117689, March 2021.
\newblock ISSN 10538119.
\newblock \doi{10.1016/j.neuroimage.2020.117689}.
\newblock URL
  \url{https://linkinghub.elsevier.com/retrieve/pii/S1053811920311745}.

\bibitem[Egger et~al.(2017)Egger, Opfer, Wang, Kepp, Sormani, Spies, Barnett,
  and Schippling]{egger_mri_2017}
Christine Egger, Roland Opfer, Chenyu Wang, Timo Kepp, Maria~Pia Sormani,
  Lothar Spies, Michael Barnett, and Sven Schippling.
\newblock {MRI} {FLAIR} lesion segmentation in multiple sclerosis: {Does}
  automated segmentation hold up with manual annotation?
\newblock \emph{NeuroImage: Clinical}, 13:\penalty0 264--270, 2017.

\bibitem[Everitt and Skrondal(2002)]{everitt_cambridge_2002}
Brian Everitt and Anders Skrondal.
\newblock \emph{The {Cambridge} dictionary of statistics}, volume 106.
\newblock Cambridge university press Cambridge, 2002.

\bibitem[Frid-Adar et~al.(2018)Frid-Adar, Klang, Amitai, Goldberger, and
  Greenspan]{frid-adar_synthetic_2018}
Maayan Frid-Adar, Eyal Klang, Michal Amitai, Jacob Goldberger, and Hayit
  Greenspan.
\newblock Synthetic {Data} {Augmentation} using {GAN} for {Improved} {Liver}
  {Lesion} {Classification}.
\newblock \emph{arXiv:1801.02385 [cs]}, January 2018.
\newblock URL \url{http://arxiv.org/abs/1801.02385}.
\newblock arXiv: 1801.02385.

\bibitem[Ganin et~al.(2016)Ganin, Ustinova, Ajakan, Germain, Larochelle,
  Laviolette, Marchand, and Lempitsky]{ganin_domain-adversarial_2016}
Yaroslav Ganin, Evgeniya Ustinova, Hana Ajakan, Pascal Germain, Hugo
  Larochelle, François Laviolette, Mario Marchand, and Victor Lempitsky.
\newblock Domain-{Adversarial} {Training} of {Neural} {Networks}.
\newblock \emph{arXiv:1505.07818 [cs, stat]}, May 2016.
\newblock URL \url{http://arxiv.org/abs/1505.07818}.
\newblock arXiv: 1505.07818.

\bibitem[Ghafoorian et~al.(2017)Ghafoorian, Mehrtash, Kapur, Karssemeijer,
  Marchiori, Pesteie, Guttmann, de~Leeuw, Tempany, van Ginneken, Fedorov,
  Abolmaesumi, Platel, and Wells~III]{ghafoorian_transfer_2017}
Mohsen Ghafoorian, Alireza Mehrtash, Tina Kapur, Nico Karssemeijer, Elena
  Marchiori, Mehran Pesteie, Charles R.~G. Guttmann, Frank-Erik de~Leeuw,
  Clare~M. Tempany, Bram van Ginneken, Andriy Fedorov, Purang Abolmaesumi, Bram
  Platel, and William~M. Wells~III.
\newblock Transfer {Learning} for {Domain} {Adaptation} in {MRI}: {Application}
  in {Brain} {Lesion} {Segmentation}.
\newblock \emph{arXiv:1702.07841 [cs]}, 10435:\penalty0 516--524, 2017.
\newblock \doi{10.1007/978-3-319-66179-7_59}.
\newblock URL \url{http://arxiv.org/abs/1702.07841}.
\newblock arXiv: 1702.07841.

\bibitem[Goodfellow et~al.(2014)Goodfellow, Pouget-Abadie, Mirza, Xu,
  Warde-Farley, Ozair, Courville, and Bengio]{goodfellow_generative_2014}
Ian Goodfellow, Jean Pouget-Abadie, Mehdi Mirza, Bing Xu, David Warde-Farley,
  Sherjil Ozair, Aaron Courville, and Yoshua Bengio.
\newblock Generative adversarial nets.
\newblock In \emph{Advances in neural information processing systems}, pages
  2672--2680, 2014.

\bibitem[Gretton et~al.(2008)Gretton, Smola, Huang, Schmittfull, Borgwardt, and
  Schölkopf]{quinonero-candela_covariate_2008}
Arthur Gretton, Alex Smola, Jiayuan Huang, Marcel Schmittfull, Karsten
  Borgwardt, and Bernhard Schölkopf.
\newblock Covariate {Shift} by {Kernel} {Mean} {Matching}.
\newblock In Joaquin Quiñonero-Candela, Masashi Sugiyama, Anton Schwaighofer,
  and Neil~D. Lawrence, editors, \emph{Dataset {Shift} in {Machine}
  {Learning}}, pages 131--160. The MIT Press, December 2008.
\newblock ISBN 978-0-262-17005-5.
\newblock \doi{10.7551/mitpress/9780262170055.003.0008}.
\newblock URL
  \url{http://mitpress.universitypressscholarship.com/view/10.7551/mitpress/9780262170055.001.0001/upso-9780262170055-chapter-8}.

\bibitem[He et~al.(2015)He, Zhang, Ren, and Sun]{he_deep_2015}
Kaiming He, Xiangyu Zhang, Shaoqing Ren, and Jian Sun.
\newblock Deep {Residual} {Learning} for {Image} {Recognition}.
\newblock \emph{arXiv:1512.03385 [cs]}, December 2015.
\newblock URL \url{http://arxiv.org/abs/1512.03385}.
\newblock arXiv: 1512.03385.

\bibitem[Heusel et~al.(2018)Heusel, Ramsauer, Unterthiner, Nessler, and
  Hochreiter]{heusel_gans_2018}
Martin Heusel, Hubert Ramsauer, Thomas Unterthiner, Bernhard Nessler, and Sepp
  Hochreiter.
\newblock {GANs} {Trained} by a {Two} {Time}-{Scale} {Update} {Rule} {Converge}
  to a {Local} {Nash} {Equilibrium}.
\newblock \emph{arXiv:1706.08500 [cs, stat]}, January 2018.
\newblock URL \url{http://arxiv.org/abs/1706.08500}.
\newblock arXiv: 1706.08500.

\bibitem[Hung et~al.(2018)Hung, Tsai, Liou, Lin, and
  Yang]{hung_adversarial_2018}
Wei-Chih Hung, Yi-Hsuan Tsai, Yan-Ting Liou, Yen-Yu Lin, and Ming-Hsuan Yang.
\newblock Adversarial {Learning} for {Semi}-{Supervised} {Semantic}
  {Segmentation}.
\newblock \emph{arXiv:1802.07934 [cs]}, July 2018.
\newblock URL \url{http://arxiv.org/abs/1802.07934}.
\newblock arXiv: 1802.07934.

\bibitem[Huo et~al.(2019)Huo, Xu, Moon, Bao, Assad, Moyo, Savona, Abramson, and
  Landman]{huo_synseg-net_2019}
Yuankai Huo, Zhoubing Xu, Hyeonsoo Moon, Shunxing Bao, Albert Assad, Tamara~K.
  Moyo, Michael~R. Savona, Richard~G. Abramson, and Bennett~A. Landman.
\newblock {SynSeg}-{Net}: {Synthetic} {Segmentation} {Without} {Target}
  {Modality} {Ground} {Truth}.
\newblock \emph{IEEE Transactions on Medical Imaging}, 38\penalty0
  (4):\penalty0 1016--1025, April 2019.
\newblock ISSN 1558-254X.
\newblock \doi{10.1109/TMI.2018.2876633}.
\newblock Conference Name: IEEE Transactions on Medical Imaging.

\bibitem[Isola et~al.(2018)Isola, Zhu, Zhou, and
  Efros]{isola_image--image_2018}
Phillip Isola, Jun-Yan Zhu, Tinghui Zhou, and Alexei~A. Efros.
\newblock Image-to-{Image} {Translation} with {Conditional} {Adversarial}
  {Networks}.
\newblock \emph{arXiv:1611.07004 [cs]}, November 2018.
\newblock URL \url{http://arxiv.org/abs/1611.07004}.
\newblock arXiv: 1611.07004.

\bibitem[Jiang et~al.(2018)Jiang, Hu, Tyagi, Zhang, Rimner, Mageras, Deasy, and
  Veeraraghavan]{jiang_tumor-aware_2018}
Jue Jiang, Yu-Chi Hu, Neelam Tyagi, Pengpeng Zhang, Andreas Rimner, Gig~S.
  Mageras, Joseph~O. Deasy, and Harini Veeraraghavan.
\newblock Tumor-aware, {Adversarial} {Domain} {Adaptation} from {CT} to {MRI}
  for {Lung} {Cancer} {Segmentation}.
\newblock \emph{Medical image computing and computer-assisted intervention :
  MICCAI ... International Conference on Medical Image Computing and
  Computer-Assisted Intervention}, 11071:\penalty0 777--785, September 2018.
\newblock \doi{10.1007/978-3-030-00934-2_86}.
\newblock URL \url{https://www.ncbi.nlm.nih.gov/pmc/articles/PMC6169798/}.

\bibitem[Jolliffe and Cadima(2016)]{jolliffe_principal_2016}
Ian~T. Jolliffe and Jorge Cadima.
\newblock Principal component analysis: a review and recent developments.
\newblock \emph{Philosophical Transactions of the Royal Society A:
  Mathematical, Physical and Engineering Sciences}, 374\penalty0
  (2065):\penalty0 20150202, April 2016.
\newblock ISSN 1364-503X, 1471-2962.
\newblock \doi{10.1098/rsta.2015.0202}.
\newblock URL
  \url{https://royalsocietypublishing.org/doi/10.1098/rsta.2015.0202}.

\bibitem[Karras et~al.(2019)Karras, Laine, and Aila]{karras_style-based_2019}
Tero Karras, Samuli Laine, and Timo Aila.
\newblock A {Style}-{Based} {Generator} {Architecture} for {Generative}
  {Adversarial} {Networks}.
\newblock \emph{arXiv:1812.04948 [cs, stat]}, March 2019.
\newblock URL \url{http://arxiv.org/abs/1812.04948}.
\newblock arXiv: 1812.04948.

\bibitem[Karras et~al.(2020)Karras, Aittala, Hellsten, Laine, Lehtinen, and
  Aila]{karras_training_2020}
Tero Karras, Miika Aittala, Janne Hellsten, Samuli Laine, Jaakko Lehtinen, and
  Timo Aila.
\newblock Training {Generative} {Adversarial} {Networks} with {Limited} {Data}.
\newblock \emph{arXiv:2006.06676 [cs, stat]}, October 2020.
\newblock URL \url{http://arxiv.org/abs/2006.06676}.
\newblock arXiv: 2006.06676.

\bibitem[Khademi et~al.(2021)Khademi, Gibicar, Arezza, DiGregorio, Tyrrell, and
  Moody]{khademi_segmentation_2021}
April Khademi, Adam Gibicar, Giordano Arezza, Justin DiGregorio, Pascal~N.
  Tyrrell, and Alan~R. Moody.
\newblock Segmentation of white matter lesions in multicentre {FLAIR} {MRI}.
\newblock \emph{Neuroimage: Reports}, 1\penalty0 (4):\penalty0 100044, December
  2021.
\newblock ISSN 26669560.
\newblock \doi{10.1016/j.ynirp.2021.100044}.
\newblock URL
  \url{https://linkinghub.elsevier.com/retrieve/pii/S2666956021000428}.

\bibitem[Khoreva et~al.(2016)Khoreva, Benenson, Hosang, Hein, and
  Schiele]{khoreva_simple_2016}
Anna Khoreva, Rodrigo Benenson, Jan Hosang, Matthias Hein, and Bernt Schiele.
\newblock Simple {Does} {It}: {Weakly} {Supervised} {Instance} and {Semantic}
  {Segmentation}.
\newblock \emph{arXiv:1603.07485 [cs]}, November 2016.
\newblock URL \url{http://arxiv.org/abs/1603.07485}.
\newblock arXiv: 1603.07485.

\bibitem[Kingma and Ba(2017)]{kingma_adam_2017}
Diederik~P. Kingma and Jimmy Ba.
\newblock Adam: {A} {Method} for {Stochastic} {Optimization}.
\newblock \emph{arXiv:1412.6980 [cs]}, January 2017.
\newblock URL \url{http://arxiv.org/abs/1412.6980}.
\newblock arXiv: 1412.6980.

\bibitem[Kuijf et~al.(2019)Kuijf, Biesbroek, de~Bresser, Heinen, Andermatt,
  Bento, Berseth, Belyaev, Cardoso, Casamitjana, Collins, Dadar, Georgiou,
  Ghafoorian, Jin, Khademi, Knight, Li, Lladó, Luna, Mahmood, McKinley,
  Mehrtash, Ourselin, Park, Park, Park, Pezold, Puybareau, Rittner, Sudre,
  Valverde, Vilaplana, Wiest, Xu, Xu, Zeng, Zhang, Zheng, Chen, van~der Flier,
  Barkhof, Viergever, and Biessels]{kuijf_standardized_2019}
Hugo~J. Kuijf, J.~Matthijs Biesbroek, Jeroen de~Bresser, Rutger Heinen, Simon
  Andermatt, Mariana Bento, Matt Berseth, Mikhail Belyaev, M.~Jorge Cardoso,
  Adrià Casamitjana, D.~Louis Collins, Mahsa Dadar, Achilleas Georgiou, Mohsen
  Ghafoorian, Dakai Jin, April Khademi, Jesse Knight, Hongwei Li, Xavier
  Lladó, Miguel Luna, Qaiser Mahmood, Richard McKinley, Alireza Mehrtash,
  Sébastien Ourselin, Bo-yong Park, Hyunjin Park, Sang~Hyun Park, Simon
  Pezold, Elodie Puybareau, Leticia Rittner, Carole~H. Sudre, Sergi Valverde,
  Verónica Vilaplana, Roland Wiest, Yongchao Xu, Ziyue Xu, Guodong Zeng,
  Jianguo Zhang, Guoyan Zheng, Christopher Chen, Wiesje van~der Flier, Frederik
  Barkhof, Max~A. Viergever, and Geert~Jan Biessels.
\newblock Standardized {Assessment} of {Automatic} {Segmentation} of {White}
  {Matter} {Hyperintensities} and {Results} of the {WMH} {Segmentation}
  {Challenge}.
\newblock \emph{IEEE Transactions on Medical Imaging}, 38\penalty0
  (11):\penalty0 2556--2568, November 2019.
\newblock ISSN 0278-0062, 1558-254X.
\newblock \doi{10.1109/TMI.2019.2905770}.
\newblock URL \url{http://arxiv.org/abs/1904.00682}.
\newblock arXiv: 1904.00682.

\bibitem[Li et~al.(2021)Li, Yang, Kreis, Torralba, and
  Fidler]{li_semantic_2021}
Daiqing Li, Junlin Yang, Karsten Kreis, Antonio Torralba, and Sanja Fidler.
\newblock Semantic {Segmentation} with {Generative} {Models}:
  {Semi}-{Supervised} {Learning} and {Strong} {Out}-of-{Domain}
  {Generalization}.
\newblock \emph{arXiv:2104.05833 [cs]}, April 2021.
\newblock URL \url{http://arxiv.org/abs/2104.05833}.
\newblock arXiv: 2104.05833.

\bibitem[Long et~al.(2015)Long, Shelhamer, and Darrell]{long_fully_2015}
Jonathan Long, Evan Shelhamer, and Trevor Darrell.
\newblock Fully {Convolutional} {Networks} for {Semantic} {Segmentation}.
\newblock \emph{arXiv:1411.4038 [cs]}, March 2015.
\newblock URL \url{http://arxiv.org/abs/1411.4038}.
\newblock arXiv: 1411.4038.

\bibitem[Lucic et~al.(2018)Lucic, Kurach, Michalski, Gelly, and
  Bousquet]{lucic_are_2018}
Mario Lucic, Karol Kurach, Marcin Michalski, Sylvain Gelly, and Olivier
  Bousquet.
\newblock Are {GANs} {Created} {Equal}? {A} {Large}-{Scale} {Study}.
\newblock \emph{arXiv:1711.10337 [cs, stat]}, October 2018.
\newblock URL \url{http://arxiv.org/abs/1711.10337}.
\newblock arXiv: 1711.10337.

\bibitem[Palladino et~al.(2020)Palladino, Slezak, and
  Ferrante]{palladino_unsupervised_2020}
Julian~Alberto Palladino, Diego~Fernandez Slezak, and Enzo Ferrante.
\newblock Unsupervised {Domain} {Adaptation} via {CycleGAN} for {White}
  {Matter} {Hyperintensity} {Segmentation} in {Multicenter} {MR} {Images}.
\newblock \emph{arXiv:2009.04985 [cs, eess]}, September 2020.
\newblock URL \url{http://arxiv.org/abs/2009.04985}.
\newblock arXiv: 2009.04985.

\bibitem[Polman et~al.(2011)Polman, Reingold, Banwell, Clanet, Cohen, Filippi,
  Fujihara, Havrdova, Hutchinson, Kappos, and {others}]{polman_diagnostic_2011}
Chris~H Polman, Stephen~C Reingold, Brenda Banwell, Michel Clanet, Jeffrey~A
  Cohen, Massimo Filippi, Kazuo Fujihara, Eva Havrdova, Michael Hutchinson,
  Ludwig Kappos, and {others}.
\newblock Diagnostic criteria for multiple sclerosis: 2010 revisions to the
  {McDonald} criteria.
\newblock \emph{Annals of neurology}, 69\penalty0 (2):\penalty0 292--302, 2011.

\bibitem[Qi et~al.(2016)Qi, Liu, Shi, Zhao, and Jia]{leibe_augmented_2016}
Xiaojuan Qi, Zhengzhe Liu, Jianping Shi, Hengshuang Zhao, and Jiaya Jia.
\newblock Augmented {Feedback} in {Semantic} {Segmentation} {Under} {Image}
  {Level} {Supervision}.
\newblock In Bastian Leibe, Jiri Matas, Nicu Sebe, and Max Welling, editors,
  \emph{Computer {Vision} – {ECCV} 2016}, volume 9912, pages 90--105.
  Springer International Publishing, Cham, 2016.
\newblock ISBN 978-3-319-46483-1 978-3-319-46484-8.
\newblock \doi{10.1007/978-3-319-46484-8_6}.
\newblock URL \url{http://link.springer.com/10.1007/978-3-319-46484-8_6}.
\newblock Series Title: Lecture Notes in Computer Science.

\bibitem[Radford et~al.(2016)Radford, Metz, and
  Chintala]{radford_unsupervised_2016}
Alec Radford, Luke Metz, and Soumith Chintala.
\newblock Unsupervised {Representation} {Learning} with {Deep} {Convolutional}
  {Generative} {Adversarial} {Networks}.
\newblock \emph{arXiv:1511.06434 [cs]}, January 2016.
\newblock URL \url{http://arxiv.org/abs/1511.06434}.
\newblock arXiv: 1511.06434.

\bibitem[Reiche et~al.(2019)Reiche, Moody, and
  Khademi]{reiche_pathology-preserving_2019}
Brittany Reiche, A.R. Moody, and April Khademi.
\newblock Pathology-preserving intensity standardization framework for
  multi-institutional {FLAIR} {MRI} datasets.
\newblock \emph{Magnetic Resonance Imaging}, 62:\penalty0 59--69, October 2019.
\newblock ISSN 0730725X.
\newblock \doi{10.1016/j.mri.2019.05.001}.
\newblock URL
  \url{https://linkinghub.elsevier.com/retrieve/pii/S0730725X18304120}.

\bibitem[Ronneberger et~al.(2015)Ronneberger, Fischer, and
  Brox]{ronneberger_u-net_2015}
Olaf Ronneberger, Philipp Fischer, and Thomas Brox.
\newblock U-{Net}: {Convolutional} {Networks} for {Biomedical} {Image}
  {Segmentation}.
\newblock \emph{arXiv:1505.04597 [cs]}, May 2015.
\newblock URL \url{http://arxiv.org/abs/1505.04597}.
\newblock arXiv: 1505.04597.

\bibitem[Sandfort et~al.(2019)Sandfort, Yan, Pickhardt, and
  Summers]{sandfort_data_2019}
Veit Sandfort, Ke~Yan, Perry~J. Pickhardt, and Ronald~M. Summers.
\newblock Data augmentation using generative adversarial networks ({CycleGAN})
  to improve generalizability in {CT} segmentation tasks.
\newblock \emph{Scientific Reports}, 9\penalty0 (1):\penalty0 16884, December
  2019.
\newblock ISSN 2045-2322.
\newblock \doi{10.1038/s41598-019-52737-x}.
\newblock URL \url{http://www.nature.com/articles/s41598-019-52737-x}.

\bibitem[Seitzer(2020)]{seitzer_pytorch-fid_2020}
Maximilian Seitzer.
\newblock pytorch-fid: {FID} {Score} for {PyTorch}, August 2020.
\newblock URL \url{https://github.com/mseitzer/pytorch-fid}.

\bibitem[Shrivastava et~al.(2017)Shrivastava, Pfister, Tuzel, Susskind, Wang,
  and Webb]{shrivastava_learning_2017}
Ashish Shrivastava, Tomas Pfister, Oncel Tuzel, Josh Susskind, Wenda Wang, and
  Russ Webb.
\newblock Learning from {Simulated} and {Unsupervised} {Images} through
  {Adversarial} {Training}.
\newblock \emph{arXiv:1612.07828 [cs]}, July 2017.
\newblock URL \url{http://arxiv.org/abs/1612.07828}.
\newblock arXiv: 1612.07828.

\bibitem[Steenwijk et~al.(2013)Steenwijk, Pouwels, Daams, van Dalen, Caan,
  Richard, Barkhof, and Vrenken]{steenwijk_accurate_2013}
Martijn~D. Steenwijk, Petra~J.W. Pouwels, Marita Daams, Jan~Willem van Dalen,
  Matthan~W.A. Caan, Edo Richard, Frederik Barkhof, and Hugo Vrenken.
\newblock Accurate white matter lesion segmentation by k nearest neighbor
  classification with tissue type priors ({kNN}-{TTPs}).
\newblock \emph{NeuroImage: Clinical}, 3:\penalty0 462--469, 2013.
\newblock ISSN 22131582.
\newblock \doi{10.1016/j.nicl.2013.10.003}.
\newblock URL
  \url{https://linkinghub.elsevier.com/retrieve/pii/S2213158213001332}.

\bibitem[Sudre et~al.(2017)Sudre, Li, Vercauteren, Ourselin, and
  Cardoso]{sudre_generalised_2017}
Carole~H. Sudre, Wenqi Li, Tom Vercauteren, Sébastien Ourselin, and M.~Jorge
  Cardoso.
\newblock Generalised {Dice} overlap as a deep learning loss function for
  highly unbalanced segmentations.
\newblock \emph{arXiv:1707.03237 [cs]}, 10553:\penalty0 240--248, 2017.
\newblock \doi{10.1007/978-3-319-67558-9_28}.
\newblock URL \url{http://arxiv.org/abs/1707.03237}.
\newblock arXiv: 1707.03237.

\bibitem[Sundaresan et~al.(2021)Sundaresan, Zamboni, Dinsdale, Rothwell,
  Griffanti, and Jenkinson]{sundaresan_comparison_2021}
Vaanathi Sundaresan, Giovanna Zamboni, Nicola~K. Dinsdale, Peter~M. Rothwell,
  Ludovica Griffanti, and Mark Jenkinson.
\newblock Comparison of domain adaptation techniques for white matter
  hyperintensity segmentation in brain {MR} images.
\newblock preprint, Neuroscience, March 2021.
\newblock URL \url{http://biorxiv.org/lookup/doi/10.1101/2021.03.12.435171}.

\bibitem[Tardif et~al.(2013)Tardif, Spence, Heinonen, Moody, Pressacco, Frayne,
  L'Allier, Chow, Friedrich, Black, Fenster, Rutt, and
  Beanlands]{tardif_atherosclerosis_2013}
Jean-Claude Tardif, J.~David Spence, Therese~M. Heinonen, Alan Moody, Josephine
  Pressacco, Richard Frayne, Philippe L'Allier, Benjamin~J.W. Chow, Matthias
  Friedrich, Sandra~E. Black, Aaron Fenster, Brian Rutt, and Rob Beanlands.
\newblock Atherosclerosis {Imaging} and the {Canadian} {Atherosclerosis}
  {Imaging} {Network}.
\newblock \emph{Canadian Journal of Cardiology}, 29\penalty0 (3):\penalty0
  297--303, March 2013.
\newblock ISSN 0828282X.
\newblock \doi{10.1016/j.cjca.2012.09.017}.
\newblock URL
  \url{https://linkinghub.elsevier.com/retrieve/pii/S0828282X12013694}.

\bibitem[Van~der Maaten and Hinton(2008)]{van_der_maaten_visualizing_2008}
Laurens Van~der Maaten and Geoffrey Hinton.
\newblock Visualizing data using t-{SNE}.
\newblock \emph{Journal of machine learning research}, 9\penalty0 (11), 2008.

\bibitem[Van~der Walt et~al.(2014)Van~der Walt, Schönberger, Nunez-Iglesias,
  Boulogne, Warner, Yager, Gouillart, and Yu]{van_der_walt_scikit-image_2014}
Stefan Van~der Walt, Johannes~L Schönberger, Juan Nunez-Iglesias, François
  Boulogne, Joshua~D Warner, Neil Yager, Emmanuelle Gouillart, and Tony Yu.
\newblock scikit-image: image processing in {Python}.
\newblock \emph{PeerJ}, 2:\penalty0 e453, 2014.

\bibitem[Volpi et~al.(2018)Volpi, Namkoong, Sener, Duchi, Murino, and
  Savarese]{volpi_generalizing_2018}
Riccardo Volpi, Hongseok Namkoong, Ozan Sener, John Duchi, Vittorio Murino, and
  Silvio Savarese.
\newblock Generalizing to unseen domains via adversarial data augmentation.
\newblock \emph{arXiv preprint arXiv:1805.12018}, 2018.

\bibitem[Wardlaw et~al.(2015)Wardlaw, Valdés~Hernández, and
  Muñoz-Maniega]{wardlaw_what_2015}
Joanna~M Wardlaw, Maria~C Valdés~Hernández, and Susana Muñoz-Maniega.
\newblock What are {White} {Matter} {Hyperintensities} {Made} of?
\newblock \emph{Journal of the American Heart Association: Cardiovascular and
  Cerebrovascular Disease}, 4\penalty0 (6):\penalty0 e001140, June 2015.
\newblock ISSN 2047-9980.
\newblock \doi{10.1161/JAHA.114.001140}.
\newblock URL \url{https://www.ncbi.nlm.nih.gov/pmc/articles/PMC4599520/}.

\bibitem[Wilson and Cook(2020)]{wilson_survey_2020}
Garrett Wilson and Diane~J. Cook.
\newblock A {Survey} of {Unsupervised} {Deep} {Domain} {Adaptation}.
\newblock \emph{arXiv:1812.02849 [cs, stat]}, February 2020.
\newblock URL \url{http://arxiv.org/abs/1812.02849}.
\newblock arXiv: 1812.02849.

\bibitem[Wolterink et~al.(2017)Wolterink, Dinkla, Savenije, Seevinck, Berg, and
  Isgum]{wolterink_deep_2017}
Jelmer~M. Wolterink, Anna~M. Dinkla, Mark H.~F. Savenije, Peter~R. Seevinck,
  Cornelis A. T. van~den Berg, and Ivana Isgum.
\newblock Deep {MR} to {CT} {Synthesis} using {Unpaired} {Data}.
\newblock \emph{arXiv:1708.01155 [cs]}, August 2017.
\newblock URL \url{http://arxiv.org/abs/1708.01155}.
\newblock arXiv: 1708.01155.

\bibitem[You et~al.(2019)You, Long, Wang, and Jordan]{you_how_2019}
Kaichao You, Mingsheng Long, Jianmin Wang, and Michael~I. Jordan.
\newblock How {Does} {Learning} {Rate} {Decay} {Help} {Modern} {Neural}
  {Networks}?
\newblock \emph{arXiv:1908.01878 [cs, stat]}, September 2019.
\newblock URL \url{http://arxiv.org/abs/1908.01878}.
\newblock arXiv: 1908.01878.

\bibitem[Zenisek et~al.(2019)Zenisek, Holzinger, and
  Affenzeller]{zenisek_machine_2019}
Jan Zenisek, Florian Holzinger, and Michael Affenzeller.
\newblock Machine learning based concept drift detection for predictive
  maintenance.
\newblock \emph{Computers \& Industrial Engineering}, 137:\penalty0 106031,
  November 2019.
\newblock ISSN 03608352.
\newblock \doi{10.1016/j.cie.2019.106031}.
\newblock URL
  \url{https://linkinghub.elsevier.com/retrieve/pii/S0360835219304905}.

\bibitem[Zhang et~al.(2021)Zhang, Ling, Gao, Yin, Lafleche, Barriuso, Torralba,
  and Fidler]{zhang_datasetgan_2021}
Yuxuan Zhang, Huan Ling, Jun Gao, Kangxue Yin, Jean-Francois Lafleche, Adela
  Barriuso, Antonio Torralba, and Sanja Fidler.
\newblock {DatasetGAN}: {Efficient} {Labeled} {Data} {Factory} with {Minimal}
  {Human} {Effort}.
\newblock \emph{arXiv:2104.06490 [cs]}, April 2021.
\newblock URL \url{http://arxiv.org/abs/2104.06490}.
\newblock arXiv: 2104.06490.

\bibitem[Zhang et~al.(2019)Zhang, Yang, and Zheng]{zhang_translating_2019}
Zizhao Zhang, Lin Yang, and Yefeng Zheng.
\newblock Translating and {Segmenting} {Multimodal} {Medical} {Volumes} with
  {Cycle}- and {Shape}-{Consistency} {Generative} {Adversarial} {Network}.
\newblock \emph{arXiv:1802.09655 [cs]}, March 2019.
\newblock URL \url{http://arxiv.org/abs/1802.09655}.
\newblock arXiv: 1802.09655.

\bibitem[Zhong et~al.(2012)Zhong, Qi, Kang, Feng, and
  Haacke]{zhong_automatic_2012}
Yi~Zhong, Shouliang Qi, Yan Kang, Wei Feng, and E.~Mark Haacke.
\newblock Automatic skull stripping in brain {MRI} based on local moment of
  inertia structure tensor.
\newblock In \emph{2012 {IEEE} {International} {Conference} on {Information}
  and {Automation}}, pages 437--440, Shenyang, China, June 2012. IEEE.
\newblock ISBN 978-1-4673-2237-9 978-1-4673-2238-6 978-1-4673-2236-2.
\newblock \doi{10.1109/ICInfA.2012.6246845}.
\newblock URL \url{http://ieeexplore.ieee.org/document/6246845/}.

\bibitem[Zhu et~al.(2020)Zhu, Park, Isola, and Efros]{zhu_unpaired_2020}
Jun-Yan Zhu, Taesung Park, Phillip Isola, and Alexei~A. Efros.
\newblock Unpaired {Image}-to-{Image} {Translation} using {Cycle}-{Consistent}
  {Adversarial} {Networks}.
\newblock \emph{arXiv:1703.10593 [cs]}, August 2020.
\newblock URL \url{http://arxiv.org/abs/1703.10593}.
\newblock arXiv: 1703.10593.

\end{thebibliography}

\end{document}